\documentclass[10pt,journal,compsoc]{IEEEtran}
\ifCLASSOPTIONcompsoc
  \usepackage[nocompress]{cite}
  \usepackage{multirow}
\else
  \usepackage{cite}
\fi
\ifCLASSINFOpdf
   \usepackage[pdftex]{graphicx}
\else
\fi
\usepackage{amsthm, amsmath}
\usepackage{amssymb}
\usepackage{mathrsfs}

\usepackage[hyphens]{url}
\usepackage{hyperref}

\usepackage{ragged2e}

\usepackage{cite}

\begin{document}
\title{Multiscale Dynamic Graph Representation for Biometric Recognition with Occlusions}

\author{Min~Ren$^{\dag}$ ,
        Yunlong~Wang$^{\dag}$ ,~\IEEEmembership{Member,~IEEE,}
        Yuhao~Zhu,
        Kunbo~Zhang,
        and~Zhenan~Sun*,~\IEEEmembership{Senior Member,~IEEE,}% <-this % stops a space
        %and~Tieniu~Tan,~\IEEEmembership{Fellow,~IEEE,}% <-this % stops a space
\IEEEcompsocitemizethanks{\IEEEcompsocthanksitem M. Ren is with the School of Artificial Intelligence, Beijing Normal University, Beijing 100875, China.\protect\\
E-mail: renmin@bnu.edu.cn
\IEEEcompsocthanksitem Y. Wang and K. Zhang are with the Center for Research on Intelligent Perception and Computing, National Laboratory of Pattern Recognition, Institute of Automation, Chinese Academy of Sciences, Beijing 100190, China.\protect\\
 E-mail: yunlong.wang@cripac.ia.ac.cn; kunbo.zhang@cripac.ia.ac.cn
\IEEEcompsocthanksitem Y. Zhu  with the Postgraduate Department, China Academy of Railway Sciences, Beijing 100081, China.\protect\\
E-mail: zhuyuhao@rails.cn
\IEEEcompsocthanksitem Z. Sun is with the Center for Research on Intelligent Perception and Computing, National Laboratory of Pattern Recognition, Institute of Automation, Chinese Academy of Sciences, Beijing 100190, China, and with CAS Center for Excellence in Brain Science and Intelligence Technology School of Artificial Intelligence, University of Chinese Academy of Sciences, Beijing 100190, China.\protect\\
 E-mail: znsun@nlpr.ia.ac.cn
 \IEEEcompsocthanksitem $\dag$ Equal Contribution
 \IEEEcompsocthanksitem * Corresponding author: Zhenan Sun.}% <-this % stops an unwanted space

}


\markboth{To be submitted to IEEE Transactions on Pattern Recognition and Machine Intelligence}%
{Shell \MakeLowercase{\textit{et al.}}: Bare Demo of IEEEtran.cls for Computer Society Journals}
\IEEEtitleabstractindextext{%
\begin{abstract}
\justifying
Occlusion is a common problem with biometric recognition in the wild.
The generalization ability of CNNs greatly decreases due to the adverse effects of various occlusions. To this end, we propose a novel unified framework integrating the merits of both CNNs and graph models to overcome occlusion problems in biometric recognition, called multiscale dynamic graph representation (MS-DGR).
More specifically, a group of deep features reflected on certain subregions is recrafted into a feature graph (FG). Each node inside the FG is deemed to characterize a specific local region of the input sample, and the edges imply the co-occurrence of non-occluded regions. 
By analyzing the similarities of the node representations and measuring the topological structures stored in the adjacent matrix, the proposed framework leverages dynamic graph matching to judiciously discard the nodes corresponding to the occluded parts. 
The multiscale strategy is further incorporated to attain more diverse nodes representing regions of various sizes.
Furthermore, the proposed framework exhibits a more illustrative and reasonable inference by showing the paired nodes.
Extensive experiments demonstrate the superiority of the proposed framework, which boosts the accuracy in both natural and occlusion-simulated cases by a large margin compared with that of baseline methods.
The source code is available \href{https://github.com/RenMin1991/Dyamic-Graph-Representation}{here}, or you can visit this website: \textit{https://github.com/RenMin1991/Dyamic-Graph-Representation}
\end{abstract}}

%\begin{IEEEkeywords}
%Computer Society, IEEE, IEEEtran, journal, \LaTeX, paper, template.
%\end{IEEEkeywords}}

\maketitle

\IEEEdisplaynontitleabstractindextext

% \fi
\IEEEpeerreviewmaketitle

\IEEEraisesectionheading{\section{Introduction}\label{sec:introduction}}

\IEEEPARstart{D}{eep} learning methods have achieved great success in recent years, especially in the area of computer vision. Convolutional neural networks (CNNs) have been widely applied as a powerful tool for image classification and feature extraction~\cite{L1998Gradient, Krizhevsky2012ImageNet, Simonyan2014Very, Szegedy2015Going, He2016Deep, Huang2017Densely, Hu2018Squeeze}. CNNs are also universally applied to biometrics, including iris recognition~\cite{Liu2016DeepIris, Zhao2017Towards, Zhang2018Deep}, face recognition~\cite{Taigman2014DeepFace, Yi2014Deep, Schroff2015FaceNet, Liu2017SphereFace, Xiang2018A, Deng2018ArcFace} and so on.

\begin{figure}[t]
\begin{center}
\includegraphics[width=0.95\linewidth]{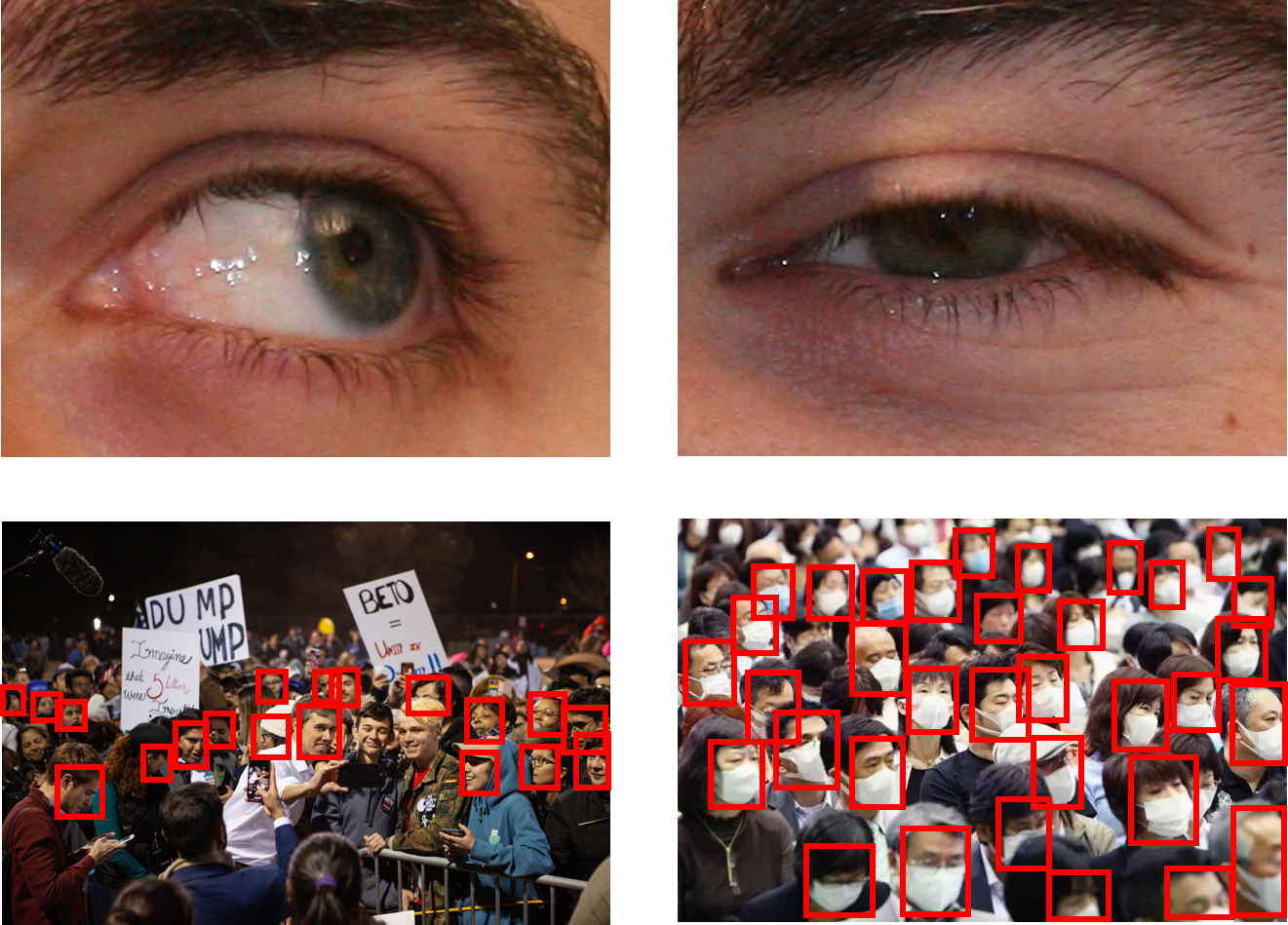}
\end{center}
\setlength{\abovecaptionskip}{0pt}
\setlength{\belowcaptionskip}{0pt}
   \caption{Iris samples and face samples in unconstrained environments. Occluded samples are common in the wild. The performance of the biometric recognition methods is greatly degraded due to the adverse effects of occlusions.}
\label{fig:occlusion}
\end{figure}

However, biometric tasks are greatly influenced by the uncertain factors of illuminations, poses, movements, facial expressions, and occlusions. Among them, occlusion is one of the most common challenges in unconstrained situations~\cite{Ban2013Face, Azeem2014A, He2018Dynamic}, as shown in Fig.~\ref{fig:occlusion}. In these situations, gaps between intraclass samples are significantly enlarged. The generalization ability of CNNs is greatly degraded due to the adverse effects of various occlusions.

To address the occlusion problem in biometric recognition, prior knowledge about occluded regions, which are usually represented as masks, is necessary for most existing biometric recognition methods~\cite{L2003, Zhenan2009Ordinal, Zhao2017Towards, Hu2013Robust, He2018Dynamic}. 
In these methods, the features of occluded areas are suppressed during matching according to the masks.
%In traditional, the masking strategy is commonly used for occlusion problems in biometrics. The features of occluded areas are suppressed during matching by masking the invalid regions out. 
However, pixel-wise labeling of masks is error-prone, and the errors are accumulated into inaccurate recognition results. More importantly, for deep learning-based methods, the features of occluded areas cannot be suppressed \emph{after} feature extraction by low-level masks in CNNs where high-level feature vectors are mostly yielded from the complete images.
It is experimentally demonstrated in Section~\ref{exp:mask} that masking the occluded areas \emph{before} feature extraction is suboptimal.

In the conference version of this paper~\cite{Ren2020Dynamic}, we proposed a novel unified framework named DGR for handling occlusion problems in biometrics. Basically, CNNs are adopted to extract deep features, and a group of deep features reflected on certain subregions are recrafted into a FG. Each node inside FG is deemed to characterize a specific local region of the input sample, and the edges imply the co-occurrence of non-occluded regions. 
%representations of local regions, graph models are then leveraged to model the relationships between these representations which act as nodes. 
%Dynamic graphs are built to remove the nodes of occluded part adaptively. 
By analyzing the similarities of the node representations and measuring the topological structures stored in the adjacent matrix, the proposed framework leverages dynamic graph matching to judiciously discard the nodes corresponding to the occluded parts. 
As a result, the similarity between intraclass samples can be significantly increased. 
%The discriminative regions and the relationships between the regions are explicitly modeled by FGs in the proposed framework.
% Modified 4-th Minor: REMOVE (good) support for decisions
In addition, the proposed framework provides the underlying reasons for the recognition decisions. The matched or non-matched nodes in the proposed dynamic graph method provide more illustrative and reasonable matching results, which serve as a support for decisions.
% Modified 4-th Minor: graph dynamic -- dynamic graph
For example, the probe face image differs from gallery images mainly in the left eye and nose regions, which can be visualized through dynamic graph matching so that it can be considered as an imposter. In other words, FG can explicitly show the reserved local regions and the similarities between these local regions. Hence, the output of the framework can be better perceived.

\begin{figure}[t]
\begin{center}
\includegraphics[width=\linewidth]{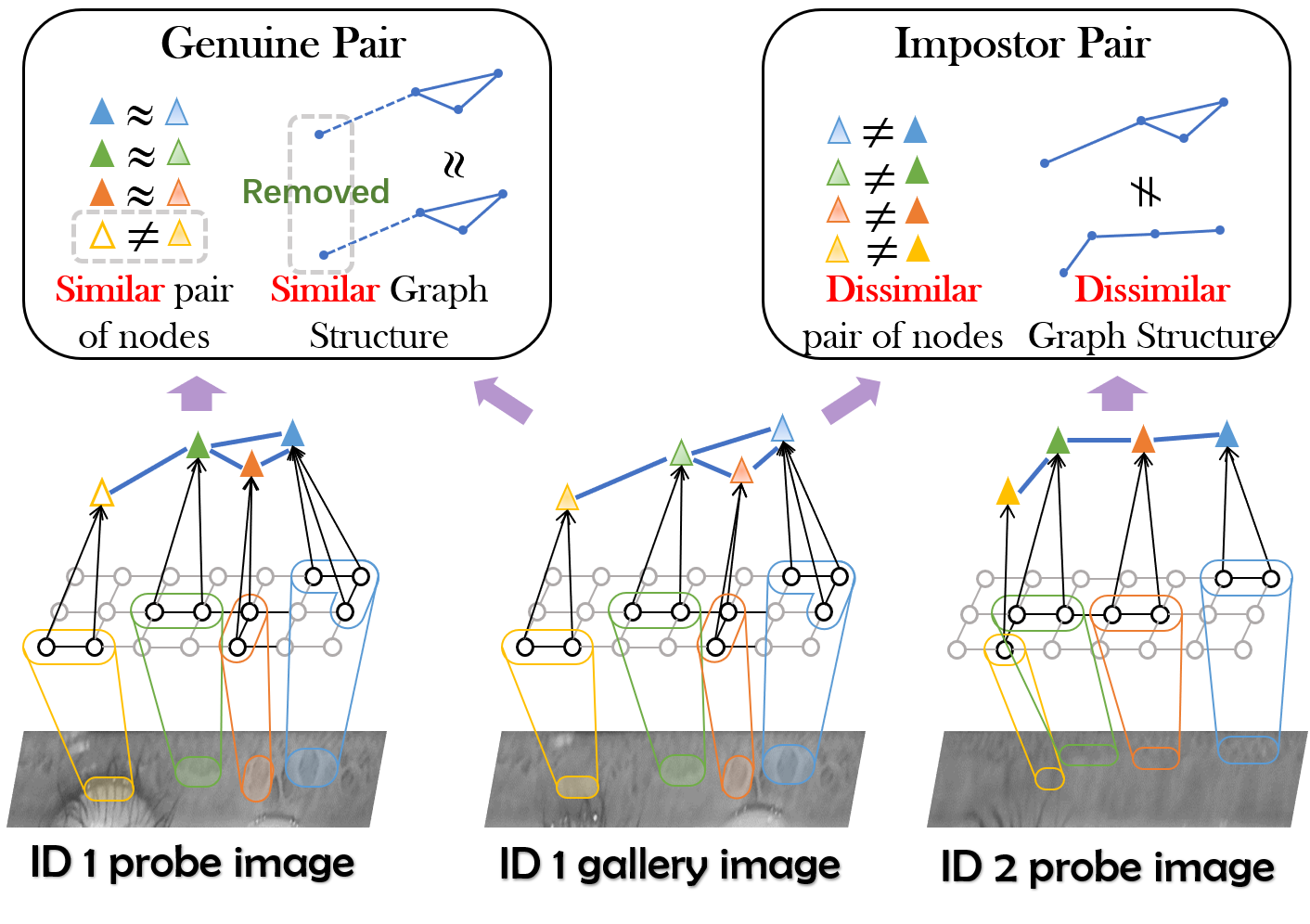}
\end{center}
\setlength{\abovecaptionskip}{0pt}
\setlength{\belowcaptionskip}{0pt}
   \caption{Illustration of our main idea. 
   A group of deep features are recrafted into a FG. The nodes of FG convey subregions of the input image, and the edges imply the relationships between these subregions. 
   To increase the intraclass similarity, the nodes and edges corresponding to the occluded parts (in the grey boxes) are automatically removed during dynamic matching.}
\label{fig:match}
\end{figure}

% Modified 4-th Minor: seamlessly combined
The graph model and CNNs are seamlessly combined: CNNs produce the feature maps for the graph generator to construct graphs, which are called FG.
The graph generator would automatically choose a subset of spatial nodes from the feature maps, convey edges to express the spatial relationships between nodes and finally assemble them together as FG.
% Modified 4-th Minor: enhance the discrimination of FG
A novel deep graph model named squeeze-and-excitation graph attention networks (SE-GAT) is proposed to enhance the discrimination of FG. 
% Modified 4-th Minor: graph dynamic -- dynamic graph
During the matching stage, dynamic graph matching is implemented. The nodes corresponding to the occluded parts are discarded accordingly by analyzing the similarities of the node representations and measuring the topological structures, as shown in Fig.~\ref{fig:match}. 
% Modified 4-th Minor: activated or not
The proposed framework is remarkably resilient to variations in the locus and ratio of occluded regions because the local features are activated or not through node selection in dynamic graph matching.

\begin{figure*}[t]
\begin{center}
\includegraphics[width=1.\linewidth]{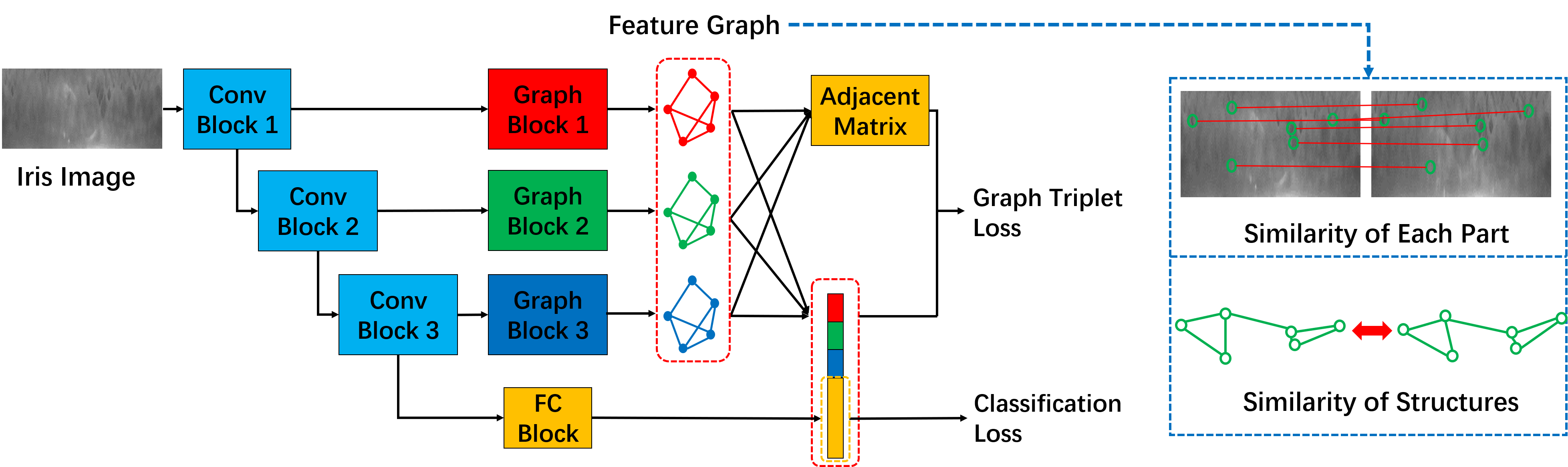}
\end{center}
 \caption{Framework of the proposed MS-DGR. Multiscale FGs are generated from feature maps of different layers. Multiscale content representations and relationship representations are contained in multiscale FGs, and all of these FGs of different scales are summarized together in the framework. The features contained in multiscale graph representation are more abundant than those in single-scale representation.}
\label{fig:framework}
\end{figure*}

This paper first extends our previous work by introducing the multiscale strategy as shown in Fig.~\ref{fig:framework}. 
The representations of nodes from one single layer can only ingest  contexts from receptive fields of the same size.
Thus, the multiscale strategy is further incorporated to attain more diverse nodes representing regions of various sizes. The primitive FGs are subsequently reorganized in a hierarchical manner for escalated dynamic graph matching.
FGs are generated from feature maps of different layers. 
% Modified 4-th Minor: and all of these FGs of different scales are summarized together in the framework.
Multiscale content representations and relationship representations are contained in multiscale FGs, and all of these FGs of different scales are summarized together in the framework.
Node features yielded from different layers correspond to different scales of local regions of the input image.
Edges from different layers represent topological structures of different scales.
Hence, the features contained in multiscale FGs representation are much more abundant than those in single-scale representation.

Second, the novel matching mechanism is of commendable scalability, which can be applied to both deep learning-based and handcrafted methods for biometric recognition.
The proposed dynamic graph matching is applied to the handcrafted features for a more convincing comparison with the masking strategy. 
More analyses and experiments are provided later to exhibit the characteristics of the proposed multiscale FGs and dynamic graph matching.

The major contributions of this paper can be summarized as follows:

\textbf{1.} We develop a novel framework for solving occlusion problems in biometrics that integrates the merits of both CNNs and graph models. The proposed framework provides a novel strategy to deal with occlusions.
%A novel deep graph model called SE-GAT is proposed as an essential component of the proposed framework.

\textbf{2.}  The proposed framework boosts the recognition performance by a large margin in comparison with baseline methods, especially in occluded scenarios.
%The decisions made by our framework are more convincing and reasonable than vanilla CNN-based methods.

\textbf{3.} The decisions made by our framework are more convincing and reasonable than vanilla CNN-based methods due to providing the underlying reasons for inference.
%similarity visualize functionality through graph dynamic matching scheme.
%The proposed framework possesses superior abilities to narrow the gaps between intra-class samples because of the dynamic graph matching, which significantly improves the generalization capability in partially occluded biometrics.

The remainder of this paper is organized as follows:
Section~\ref{sec:RelWork} presents a literature review of related works. The proposed framework is detailed in Section~\ref{sec:Method}. The configurations and results of experiments are presented in Section~\ref{sec:Experiment}. Finally, the conclusion of this paper is presented in Section~\ref{sec:Conclusion}.

%-----------------------------------------------------------------------------------
%-----------------------------------------------------------------------------------
%-----------------------------------------------------------------------------------

\section{Related Works}
\label{sec:RelWork}

In this section, we will briefly introduce the biometric approaches based on CNNs as a baseline, the handcrafted graph representation methods as references, deep graph neural networks (GNNs) as tools and the multiscale strategy in computer vision research.

Iris recognition has attracted increasing attention as one of the most accurate and reliable methods for identity authentication. Recently, CNN-based methods for iris recognition have been presented. 
The parameters of the deep learning model are tuned through data fitting. 
%Data-driven methods surpass the limitations of hand-crafted features.
DeepIris is proposed in~\cite{Liu2016DeepIris} for heterogeneous iris matching. The fully convolutional network (FCN)-based model named UniNet is proposed for iris recognition in~\cite{Zhao2017Towards}. MaxoutCNNs is proposed for iris and periocular recognition in~\cite{Zhang2018Deep}. 
However, occlusion problems, which are common for the iris, remain the bottleneck preventing improvement of the recognition performance.
Almost all existing CNN-based methods adopt the masking strategy, which is borrowed from conventional methods. However, the masking strategy is not suitable for the CNN architecture since the non-iris regions cannot be excluded after feature extraction.

The first method for face representation based on CNN was proposed by Taigman et al.~\cite{Taigman2014DeepFace}. A framework employing multiple convolutional networks was proposed by Sun et al.~\cite{Yi2014Deep}. The triplet loss function is applied to produce 128-D face embedding representations in~\cite{Schroff2015FaceNet}. A light-weight convolutional architecture is proposed for face recognition in~\cite{Xiang2018A}. 
% Modified 4-th Minor: REMOVE metric learning
Recently, several approaches for facial feature embedding enhancement have been proposed, including SphereFace~\cite{Liu2017SphereFace}, ArcFace~\cite{Deng2018ArcFace}, CosFace~\cite{Wang2018CosFace} and so on.

There are also region-based or partial-based models for face recognition~\cite{Ou2018Robust, Cheheb2017Random, He2018Dynamic, Ding2020Masked, Li2021Cropping}. 
However, similar to the masking strategy, all of these approaches need to set apart the occluded regions of face images before feature extraction.

\textbf{Graph-based methods for biometrics.} There are some existing handcrafted graph-based methods for biometrics, including iris~\cite{Kerekes2007Graphical}, face~\cite{Kisku2009Probabilistic}, periocular~\cite{Proen2013Periocular}, ear~\cite{Kisku2009Probabilistic} and hand vein~\cite{Horadam2014Hand}. Although these handcrafted methods may have inspired us, the methods cannot be directly adopted in deep learning-based frameworks.

\textbf{Deep learning approaches on graphs.} Graphs are ubiquitous in the world. 
%Deep learning methods have been extended to graph data recently~\cite{Ziwei2018Deep, Zhou2018Graph}. 
GNNs are introduced as recursive neural networks in~\cite{Gori2005A, Franco2009The} to handle graph data. Recently, the generalization of the convolutional operation has attracted increasing attention. 
Approaches in this direction can be roughly categorized as spectral approaches and nonspectral approaches~\cite{Zhou2018Graph}. The spectral approaches~\cite{Bruna2014Spectral, Kipf2016Semi} work with spectral representations of graphs. The nonspectral approaches~\cite{Duvenaud2015Convolutional, Atwood2016Diffusion, Hamilton2017Inductive} define convolutional operations directly on graphs. Graph attention networks (GAT)~\cite{Veli2017Graph} introduce the attention mechanism to deep graph models. 

Deep graph matching methods have been commonly studied recently~\cite{Zanfir2018Deep, Runzhong2019Learning, Bo2019Glmnet, Matthias2019Deep}.
Comparing to general deep learning based graph matching methods, the proposed dynamic graph matching is designed to handle graphs with dynamic nodes and edges rather than fixed structure, because the structure of FG changes according to the occluded parts. 
Besides, biometric recognition with occlusions in the real world requests to generalize on various occlusions, which is a challenge for the data-driven methods. 

\textbf{Multiscale strategy}.
The multiscale strategy is an effective approach that has been widely used in the literature. 
%Lowe proposed a famous approach for feature extraction in~\cite{Lowe1999Object} named SIFT. 
Multiscale information is utilized in SIFT~\cite{Lowe1999Object} by building an image pyramid for more stable and scale-invariant localization and description of key points.
Bay et al.~\cite{Bay2006Surf} follow the multiscale approach for key point detection and description.

The multiscale approach is also an effective strategy for deep learning frameworks.
For example, 
%in the first deep learning based semantic segmentation approach, 
information on different scales is fused together in~\cite{Long2017Fully} for deep learning-based semantic segmentation. The coarse, global-scale information from deep layers and fine, local-scale information from shallow layers are both useful.
The strategy of the multiscale pyramid is introduced for object detection in~\cite{Lin2017Feature}, where multiscale feature vectors are aggregated to process objects of different scales.

%-----------------------------------------------------------------------------------
%-----------------------------------------------------------------------------------
%-----------------------------------------------------------------------------------

\section{Multiscale Dynamic Graph Representation}
\label{sec:Method}

\begin{figure*}[t]
\begin{center}
\includegraphics[width=1.\linewidth]{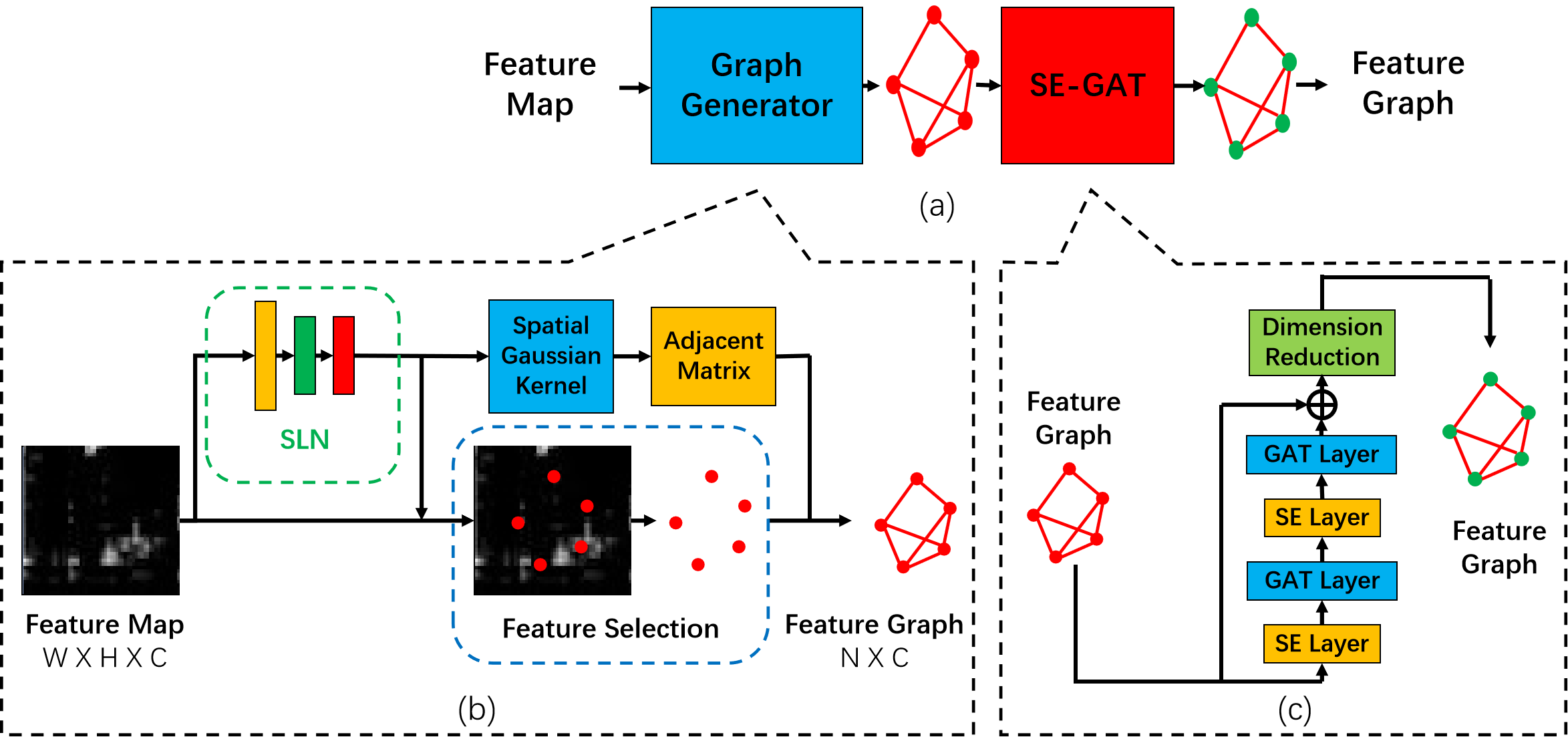}
\end{center}
%\captionsetup{justification=centering}
 \caption{Detailed structure of the graph block of Fig.~\ref{fig:framework}. \textbf{(a)} The graph block consists of a graph generator and a SE-GAT. \textbf{(b)} The feature maps from convolutional blocks are transformed to FG by the graph generator. \textbf{(c)} SE-GAT is adopted to extract high-level semantic graph representation.}
\label{fig:graph_block}
\end{figure*}

In this section, we briefly introduce the overall framework to handle the occlusion problems in biometrics. 
Two crucial components of MS-DGR include the graph generator and SE-GAT, which are elaborated separately. 
Next, a novel loss function designed for supervising the training of dynamic graph feature learning and the strategy of dynamic graph matching are introduced.

%-----------------------------------------------------------------------------------
%-----------------------------------------------------------------------------------

\subsection{Overall Framework}

In the proposed framework, input images are sent to a network that contains several convolutional blocks as shown in Fig.~\ref{fig:framework}. Each convolutional block consists of several convolutional layers and pooling layers. 
Feature maps of different scales generated by convolutional blocks are sent to graph blocks.
% Modified 4-th Minor: weighted edges
FG built upon these feature maps are generated by the graph blocks, in which nodes contain the local patterns and the weighted edges convey the topological structures of nodes.

The graph block consists of the graph generator and SE-GAT, as shown in Fig.~\ref{fig:graph_block}(a). 
% Quality control editor: Abbreviations and acronyms typically need to be defined only once within the main text. Please consider adhering to this convention.
FGs are generated from feature maps by the graph generator before being sent to SE-GAT, which is a hierarchical feature extractor based on GAT for graph-style inputs. 
FGs extracted by SE-GAT have two components: 
i. the content messages of subregions contained in the feature vectors of nodes and 
ii. the topological structure messages contained in the adjacent matrix of the graph. 
Through the multiscale strategy, nodes perceive both coarse and fine local patterns, and the adjacent matrices encode topological structures at different scales.
%FGs of different scales represent local patterns and relationships at different scale.

As shown in Fig.~\ref{fig:framework}, three graph blocks are integrated into CNNs, and the global feature is produced by the CNNs. Our framework can be incorporated into any CNNs by combining the global feature and FGs.

Finally, two kinds of loss functions are applied during training. 
The global feature is passed to a classification layer for measuring classification loss.
%The global feature generated by the fully connected layers is sent to cross-entropy loss function which measures the classification loss. 
FGs processed through SE-GAT are sent to a novel graph triplet loss function. The novel loss function provides a new way to measure the similarity of two graphs. 
Both the similarity of the node representations and the topological structures are taken into consideration.

%-----------------------------------------------------------------------------------
%-----------------------------------------------------------------------------------

\subsection{Graph Generator}

There are two steps to generate a FG from feature maps, as shown in Fig.~\ref{fig:graph_block}(b). The first step is to generate the nodes, i.e., to regress the spatial locations of nodes and selecting from the feature maps. The second step is generating the edges/adjacent matrix according to the relationships of nodes.

%\begin{figure}[h]
%\begin{center}
%\includegraphics[width=0.9\linewidth]{graph_gen.png}
%\end{center}
%   \caption{Structure of Graph Generator.}
%\label{fig:graphgen}
%\end{figure}

% Modified 4-th Minor: smaller architecture
The nodes of a FG are selected from the feature map.
Firstly, a smaller architecture is adopted to find the spatial locations of the essential features within the feature map.
The smaller architecture, which is named spatial location network (SLN), takes the feature map as input.
Its output is the spatial coordinates of the selected features.
%
% Modified 4-th Minor: 2 SENTENCE
SLN detects spatial locations of interest points through regression.
This network is trained to find the most discriminative features in the feature map.
It transforms the feature map into the spatial coordinates of nodes.
Following the conventional practice of CNNs, it consists of two convolutional layers, two pooling layers, and two fully connected layers (its layer configurations are shown in Tab.~\ref{tab:slrn}).
%
%Since the dimension of the feature map is usually quite high, the dimension is decreased gradually during the first five layers of the SLN.
%
% Modified 4-th Minor: 1 SENTENCE
The dimension of the feature map is decreased gradually during the first five layers of the SLN.
In the last fully connected layer, the feature is transformed into the spatial coordinates of nodes.
%%

%Then, the features of interest are selected according to the spatial locations.
%
% Modified 4-th Minor: 1 SENTENCE
Then, CNN features are extracted from the localized points of interest.
These selected features are the representations of nodes.
One node corresponds to one location.
The feature selection is realized by bilinear interpolation according to the spatial coordinates:
\begin{equation}
\textbf{f}_{i,j}(c) = \textbf{F}(i,j,c)
\label{equ:sample}
\end{equation}
where $\textbf{f}$ is the feature vector, $\textbf{F}$ is the feature map, $i,~j$ are the spatial coordinates which are the output of SLN, $c$ is the index of the feature vector and $c \in \{1, 2, ..., C\}$, $C$ is the number of channels of the feature map.
Now, the nodes of a FG have been generated.

%Spatial location regression of nodes is realized by a light network called the spatial location network (SLN), as shown in Fig.~\ref{fig:graph_block}(b). This network consists of two convolutional layers and two fully connected layers, which transforms a $W \times H \times C$ feature map into an $N \times 2$ coordinate matrix by regression (its layer configurations are shown in Tab.~\ref{tab:slrn}). 
%Thus, the output of SLN has $2N$ dimensions, which are normalized by the sigmoid function before multiplying by the scale of the corresponding feature map.
%The feature vector of each node is selected from the feature map by bilinear interpolation of its four neighbors according to the spatial coordinates:

%Bilinear interpolation is not shown in Eq.~\ref{equ:sample} for briefness.

\begin{table}[t]
\begin{center}
\caption{Layer configurations of SLN.}
\label{tab:slrn}
\setlength{\tabcolsep}{0.5mm}
{
\begin{tabular}{|c|c|c|c|c|c|}
\hline
\textbf{Layer}& \textbf{Kernel Size}& \textbf{Stride} & \textbf{Input Size} & \textbf{Output size} \\
\hline
Pool1 & $2\times 2$ & $2\times 2$ &$ H\times W\times C$ & $H/2\times W/2\times C$ \\
\hline
Conv1 & $5\times 5$ & $1\times 1$ & $H/2\times W/2\times C$ & $H/2\times W/2\times C/2$ \\
\hline
Pool2 & $2\times 2$ & $2\times 2$ & ~$H/2\times W/2\times C/2$~ & ~$H/4\times W/4\times C/2$~ \\
\hline
Conv2 & $5\times 5$ & $1\times 1$ & $~H/4\times W/4\times C/2$~ & ~$H/4\times W/4\times C/4$~ \\
\hline
FC1 & -- & -- & $HWC/64$ & $128$ \\
\hline
FC2 & -- & -- & $128$ & $2N$ \\
\hline
\end{tabular}}
\end{center}
\end{table}

The edges/adjacent matrix of a FG is generated according to the spatial relationships between nodes.
For two nodes of FG, the weight of the edge between them is generated by the spatial Gaussian kernel function:
\begin{equation}
\label{equ:gaussian}
\textbf{M}_{adj}(a,~b)=\left\{
\begin{array}{cl}
0 & ||\textbf{n}_a-\textbf{n}_b||_2>R, \\
exp(-\frac{||\textbf{n}_a-\textbf{n}_b||_2^2}{2 R^2}) & ||\textbf{n}_a-\textbf{n}_b||_2<R.
\end{array}
\right.
\end{equation}
where $\textbf{M}_{adj}$ is a $N\times N$ adjacent matrix, $N$ is the number of nodes, $\textbf{n}_a,~\textbf{n}_b$ are the spatial coordinates of two nodes, and $R$ is the scale of the receptive field of each node.
Hence, the edge represents the spatial relationship of the two nodes.
%
% Modified 4-th Minor: the higher the weight of the edge is
The closer the two nodes are to each other, the higher the weight of the edge is.
If the distance between the two nodes is larger than the scale of the receptive field, the weight is 0 (there is no relationship between them).
%
% Modified 4-th Minor: in this way
In this way, the edges/adjacent matrix of FG have been generated.

%%

%The reason for the design of the spatial Gaussian kernel function is that the representations of two nodes derive from shared regions of the input image if the distance between the two nodes is lower than R.
%
% Modified 4-th Minor: if their receptive fields are overlapped.
There should be an information interaction between two nodes in the graph neural network if their receptive fields are overlapped.
On the other hand, if the distance between two nodes is higher than R, their receptive fields are not overlapped, and the two nodes should not be connected.

In general, the SLN constructs the feature vectors affiliated with critical regions as nodes of FG. 
% Modified 4-th Minor: among
The spatial Gaussian kernel function establishes edges among nodes.

%------------------------------------------------------------------------
%------------------------------------------------------------------------

\subsection{Squeeze-and-Excitation Graph Attention Networks}

\begin{figure}[h]
\begin{center}
\includegraphics[width=0.85\linewidth]{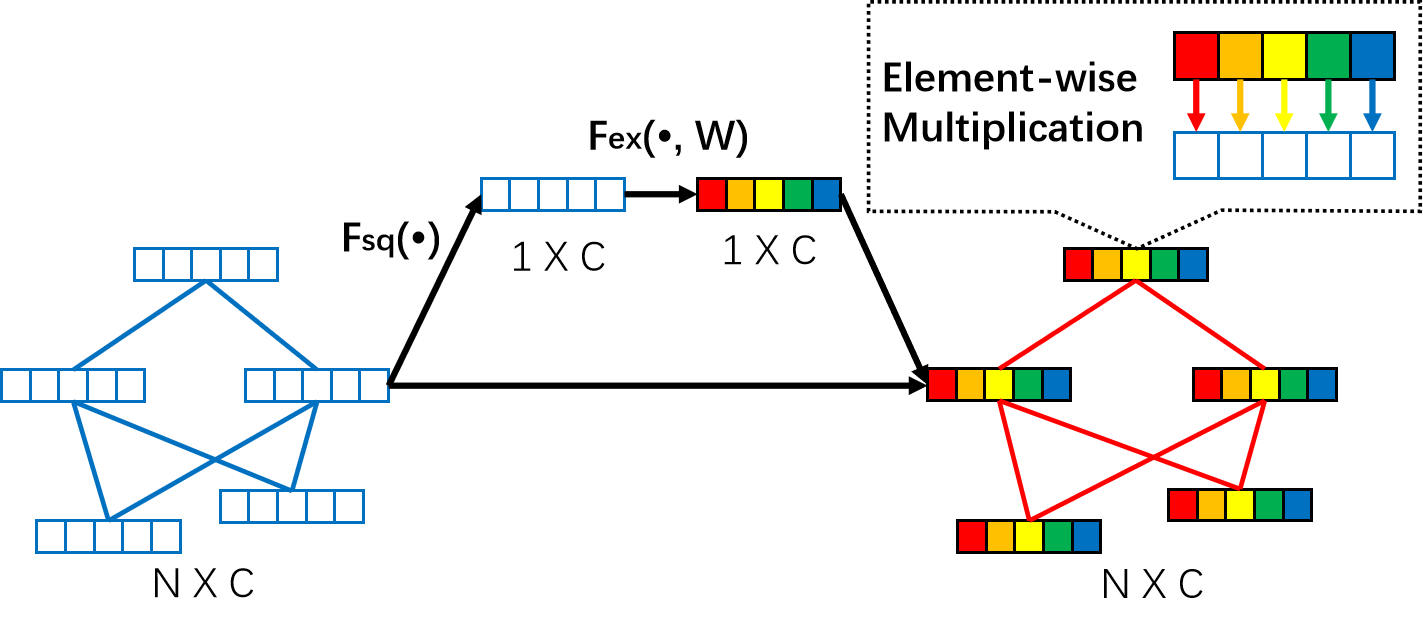}
\end{center}
   \caption{Structure of the SE layer. Different weights are assigned to channels of FG by element-wise multiplication on each node.}
\label{fig:selayer}
\end{figure}

SE-GAT is a novel model based on GAT~\cite{Veli2017Graph}. It consists of two kinds of layers: the squeeze-and-excitation (SE) layer and graph attention (GAT) layer. We describe them as follows:

The SE layer is a generalized version of the squeeze-and-excitation block~\cite{Hu2018Squeeze}, which is effective in CNNs. 
The same channel of different feature vectors represents the energy of the same pattern at different positions of the input image. The SE layer enhances the important channels and suppresses the nonsignificant channels by assigning weights.
%, which is a channel-wise attention layer, 
As shown in Fig.~\ref{fig:selayer}, the average energy of each channel of nodes is calculated by the global pooling operation:
\begin{equation}
\textbf{z}(c) = F_{sq}(\textbf{G})= \frac{1}{N}\sum_{i=1}^N \textbf{f}_i(c)
\label{equ:squeeze}
\end{equation}
where $\textbf{z}$ is the squeezed vector, $\textbf{G}$ is FG, $N$ is the number of nodes, $c$ is the index of the feature vector of each node and $c \in \{1, 2, ..., C\}$. 
Two fully connected layers are used to make use of the global information:
\begin{equation}
\textbf{s} = F_{ex}(\textbf{z},~\textbf{W}) = \delta_2(\textbf{W}_2\delta_1(\textbf{W}_1 \textbf{z})) 
\label{equ:excitation}
\end{equation}
where $\textbf{s}$ is the scale vector, $\textbf{W}_1,~\textbf{W}_2$ are parameters of the two fully connected layers, and $\delta_1,~\delta_2$ are activation functions. The scale vector $\textbf{s}$ is used to rescale each channel of FG:
\begin{equation}
\textbf{f}'_i(c) = \textbf{s}(c)\textbf{f}_i(c)
\label{equ:rescale}
\end{equation}
where $\textbf{f}'$ is the rescaled feature vector, $c$ is the channel index of the feature vector and $c \in \{1, 2, ..., C\}$, $i$ is the index of nodes and $i \in \{1, 2, ..., N\}$.

The GAT layer is a modified variant of the layer in~\cite{Veli2017Graph}. The weights of different edges of FG are different, rather than the same, in~\cite{Veli2017Graph}. The formulation of the layer is given as follows:
\begin{equation}
\textbf{f}_a' = \delta (\sum_{b\in \mathscr{N}_a} \alpha_{ab} \textbf{W} \textbf{f}_b)
\label{equ:modgat}
\end{equation}

\begin{equation}
\alpha_{ab} = \frac{exp(\delta(\textbf{M}_{adj}(a,b)e_{ab})}{ \sum_{k\in \mathscr{N}_a} exp(\delta(\textbf{M}_{adj}(a,k)e_{ak}))}
\label{equ:alpha}
\end{equation}

\begin{equation}
e_{ab} = \textbf{w}_{att}^T[\textbf{Wf}_a || \textbf{Wf}_b]
\label{equ:e}
\end{equation}
where $\textbf{f}'\in \mathbb{R}^{C}$ is the output vector, $\textbf{f}\in \mathbb{R}^{C}$ is the input vector, $\delta$ is a nonlinear activation function, $\mathscr{N}_a$ is the set of first-order neighbors of node $a$ (including $a$), $\alpha_{ab}$ is the attention value between node $a$ and node $b$, $\textbf{W}\in \mathbb{R}^{C\times C}$ and $\textbf{w}_{att}\in \mathbb{R}^{2C}$ are the parameters of the GAT layer, $\textbf{M}_{adj}$ is the adjacent matrix of the input graph, and $||$ indicates concatenation. 

%The modified GAT layer can handle more complex graphs, and the spatial relationships contained in the adjacent matrix can be exploited while the original GAT layer can not.

The module of SE-GAT is shown in Fig.~\ref{fig:graph_block}(c). As the residual structure has been proven to be effective for CNNs~\cite{He2016Deep}, we extend it to the graph model in SE-GAT. 
The residual block in the SE-GAT module has four layers, two of which are SE layers, and the rest are GAT layers.
A dimension reduction layer is adopted to reduce the dimensions of feature vectors of nodes:
\begin{equation}
\textbf{f}_i' = \delta(\textbf{W}_{C'\times C}\textbf{f}_i)
\label{equ:reduce}
\end{equation}
where $\textbf{W}_{C'\times C}$ is the parameter matrix, $\delta$ is an activation function, and
%$C'$ is the dimension of output feature vectors. 
$\textbf{f}_i'\in \mathbb{R}^{C'}$ is the output vector.
Finally, FG is extracted for matching.

%------------------------------------------------------------------------
%------------------------------------------------------------------------

\subsection{Graph Triplet Loss Function}
\label{sec:loss}

As described above, FG consists of node representations and adjacent matrices.
%The FGs contain two kinds of information: semantic information in the feature vectors and spatial relationship information in the adjacent matrices of FGs. 
Hence, the similarity between the outputs of two samples consists of two terms:
\begin{equation}
\mathit{S} = \mathit{S}_{fea} + \mathit{S}_{adj}
\label{equ:sim}
\end{equation}
where $\mathit{S}_{fea}$ is the semantic similarity and $\mathit{S}_{adj}$ is the relationship similarity. 
\begin{equation}
\mathit{S}_{fea} = cosine(\widetilde{\textbf{f}_1}, \widetilde{\textbf{f}_2}) = \frac{\widetilde{\textbf{f}_1} \cdot \widetilde{\textbf{f}_2}}{||\widetilde{\textbf{f}_1}||_2 ||\widetilde{\textbf{f}_2}||_2}
\label{equ:feasim}
\end{equation}
where $cosine$ represents the cosine similarity, and $\widetilde{\textbf{f}_1}, \widetilde{\textbf{f}_2}$ are the concatenation of all feature vectors of FGs and global features generated by fully connected layers.
\begin{equation}
\mathit{S}_{adj} = \frac{1}{N_g}\sum_{i=1}^{N_g} \frac{1}{N_i^2} ||\textbf{M}_{i}^1 - \textbf{M}_{i}^2||_F
\label{equ:adjsim}
\end{equation}
where $N_g$ is the number of FGs in a single input image extracted by the network, $N_i$ is the number of nodes of the $i$-th FG, and $\textbf{M}_{i}^1, \textbf{M}_{i}^2$ are the adjacent matrices of the $i$-th FGs of two samples.
Finally, the graph triplet loss is:
\begin{equation}
\mathit{L} = max\{0,m + S_{anc-neg} - S_{anc-pos}\}
\label{equ:graphsim}
\end{equation}
where $S_{anc-neg}$ is the similarity between anchor sample and negative sample, $S_{anc-pos}$ is the similarity between anchor sample and positive sample, and $m$ is a predefined positive margin and set to 1 during training.

%----------------------------------------------------------------------------------
%----------------------------------------------------------------------------------

\subsection{Dynamic Graph Matching}
\label{med:dgm}

Given two FGs, $G_A$ and $G_B$, from the same scale level extracted from two image samples, we firstly filter out dissimilar nodes by using the mean of cosine similarities of corresponding nodes as the threshold, which can be calculated as follows:
\begin{equation}
s_{gate} = \frac{1}{N}\sum_{i=1}^N cosine(\textbf{f}_i^A,~\textbf{f}_i^B)
\label{equ:gate}
\end{equation}
where $\textbf{f}_i^A$ and $\textbf{f}_i^B$ are the node features of $G_A$ and $G_B$, and $N$ is the number of nodes in each FG. Then, if the similarity of a node pair is less than the $s_{gate}$, this node pair is removed from the two FGs:
\begin{equation}
G_A^D, G_B^D = Re(G_A, G_B, s_{gate})
\label{equ:remove}
\end{equation}
where $Re$ represents the remove operation according to the $s_{gate}$, and the dynamic FGs $G_A^{D}$ and $G_B^{D}$ are built. Finally, multiscale dynamic FGs of two samples are built according to Eq.~\ref{equ:remove}.

%

% Modified 4-th Minor : non-occluded
The variance of occluded parts is much larger than the variance of non-occluded biometric images. The similarity between the occluded region and non-occluded region is very likely to be lower than that between non-occluded regions.
Hence, node pairs with lower similarities are removed to adaptively select features.
The threshold set by the mean of similarities provides a straightforward and effective way to build dynamic FGs.

%----------------------------------------------------------------------------------
%----------------------------------------------------------------------------------
%----------------------------------------------------------------------------------

\section{Experiments}
\label{sec:Experiment}

% Modified 4-th Minor: non-occluded
Two modalities, iris and face, are selected to evaluate the proposed framework. Experiments on non-occluded and occluded instances are both taken into consideration. Thorough experiments are launched to analyze the proposed framework, including comparison with other state-of-the-art methods in both non-occluded and occluded biometric cases, the comparison to masking strategy, the evaluation of different scales, the evaluation under different node number settings and ablation studies. In addition, the visualization of FGs further provides the underlying reasons for recognition decisions.

%----------------------------------------------------------------------------------
%----------------------------------------------------------------------------------

\begin{table}[h]
\begin{center}
\caption{Layer configurations of the network for iris recognition.}
\label{tab:irisnet}
\setlength{\tabcolsep}{0.5mm}
{
\begin{tabular}{|c|c|c|c|c|c|}
\hline
\textbf{Layer}& \textbf{Kernel Size}& \textbf{Stride} & \textbf{Padding} & \textbf{Input Size} & \textbf{Output size} \\
\hline
\hline
\multicolumn{6}{|c|}{\bf{Conv Block 1}} \\
\hline
Conv1 & $5\times 9$ & $1\times 1$ & $2\times 4$ & $128\times 256\times 1$ & $128\times 256\times 24$ \\
\hline
Pool1 & $2\times 2$ & $2\times 2$ & -- & $128\times 256\times 24$ & $64\times 128\times 24$ \\
\hline
Conv2 & $5\times 7$ & $1\times 1$ & $2\times 3$ & $64\times 128\times24$ & $64\times 128\times 48$ \\
\hline
Pool2 & $2\times 2$ & $2\times 2$ & -- &$ 64\times 128\times 48$ & $32\times 64\times 48$ \\
\hline
Conv3 & $5\times 5$ & $1\times 1$ & $2\times 2$ & $32\times 64\times48$ & $32\times 64\times 64$ \\
\hline
\multicolumn{6}{|c|}{\bf{Conv Block 2}} \\
\hline
Pool3 & $2\times 2$ & $2\times 2$ & -- &$ 32\times 64\times 64$ & $16\times 32\times 64$ \\
\hline
Conv4 & $5\times 5$ & $1\times 1$ & $2\times 2$ & $16\times 32\times64$ & $16\times 32\times 96$ \\
\hline
\multicolumn{6}{|c|}{\bf{Conv Block 3}} \\
\hline
Pool4 & $2\times 2$ & $2\times 2$ & -- & $16\times 32\times 96$ & $8\times 16\times 96$ \\
\hline
Conv5 & $5\times 5$ & $1\times 1$ & $2\times 2$ & $8\times 16\times 96$ & $8\times 16\times 96$ \\
\hline
\multicolumn{6}{|c|}{\bf{FC Block}} \\
\hline
Pool5 & $2\times 2$ & $2\times 2$ & -- &$ 8\times 16\times 96$ & $4\times 8 \times 96$ \\
\hline
FC1 & -- & -- & -- & $3072$ & $256$ \\
\hline
FC2 & -- & -- & -- & $256$ & $num\_ classes$ \\
\hline
\end{tabular}}
\end{center}
\end{table}

\subsection{Experiments of Iris Recognition}

\subsubsection{Protocols and Databases}

% Modified 4-th Minor: light-weight
A light-weight CNNs is used for iris experiments. Its layer configuration is shown in Tab.~\ref{tab:irisnet}, and the multiscale FGs consist of three scales: small scale, medium scale, and large scale. Each scale is obtained from a different convolutional block of the CNNs. 
The number of nodes decreases along with the scales of the FG. 
%Fine, local information needs more nodes to express than coarse, global information. 
The node numbers of the small-, medium-, and large-scale FG are set as 64, 32, and 16, respectively.
% Quality control editor: Please ensure that the intended meaning has been maintained in the edits of the previous sentence.
%
Stochastic gradient descent with momentum is adopted for optimization. Training is started with a learning rate of 0.001 and divided by 2 every 10 epoch and stopped at the 40th epoch. The size of the mini-batch is 64. The weight decay is 0.0001.

\begin{figure*}[h]
\begin{center}
\includegraphics[width=\linewidth]{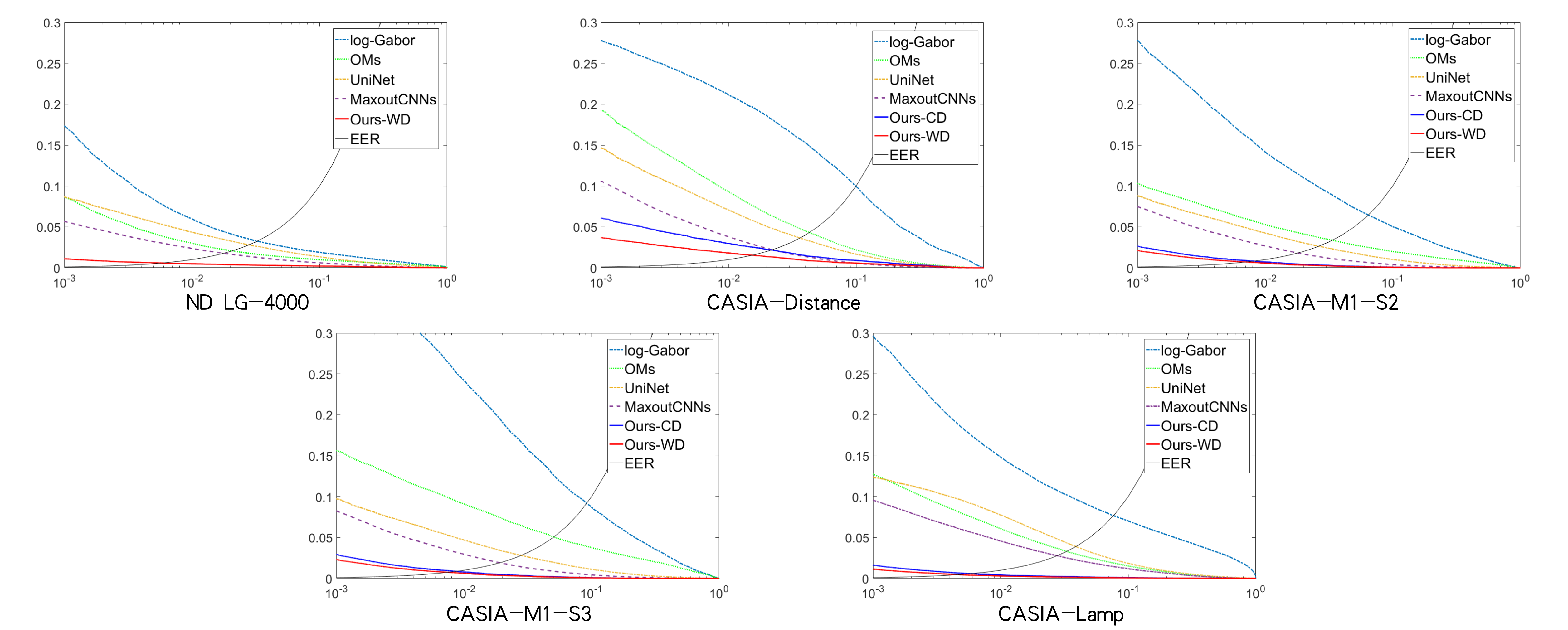}
\end{center}
   \caption{DET curves of iris recognition experiments (best viewed in color). The results show that the proposed framework is better than the compared methods on all databases. The cross-database results demonstrate the generalization ability of the proposed framework.}
\label{fig:roc_iris}
\end{figure*}

\begin{table*}[t]
\begin{center}
\caption{False rejection rates (FRR@FAR=0.01\%) and equal error rates (EER) of iris recognition experiments.}
\label{tab:iris}
\setlength{\tabcolsep}{3.5mm}
{
\begin{tabular}{c|c|c|c|c|c|c|c|c|c|c}%{p{55pt}|p{30pt}|p{25pt}|p{40pt}}%
\hline
\multirow{2}*{~} & \multicolumn{2}{|c|}{\bf{ND-LG4000}} & \multicolumn{2}{|c|}{\bf{CASIA-Distance}} & \multicolumn{2}{|c|}{\bf{CASIA-M1-S2}}& \multicolumn{2}{|c|}{\bf{CASIA-M1-S3}} & \multicolumn{2}{|c}{\bf{CASIA-Lamp}} \\
 \cline{2-11}
& FRR & EER & FRR & EER & FRR & EER & FRR & EER & FRR & EER \\
\hline
log-Gabor \cite{L2003} & 35.90\% & 3.25\% & 38.33\% & 9.98\% & 45.03\% & 6.35\%&59.84\%&9.14\% & 42.93\% & 7.63\% \\
\hline
OMs \cite{Zhenan2009Ordinal} & 17.95\% & 2.08\% & 28.88\% & 4.30\% & 16.68\% & 3.34\%&22.14\%&5.03\% & 19.30\% & 3.30\% \\
\hline
UniNet \cite{Zhao2017Towards} & 13.71\% & 2.80\% & 21.95\% & 3.65\% & 13.17\% & 2.63\%&14.58\%&2.80\% & 14.92\% & 3.89\%  \\
\hline
MaxoutCNNs \cite{Zhang2018Deep} & 9.89\% & 1.77\% & 19.61\% & 2.22\% & 13.81\% & 1.81\%&15.26\%&1.92\% & 14.48\% & 2.78\%  \\
\hline
Ours-CD & -- & -- & 9.93\% & 2.17\% & 7.33\% & 0.82\%&8.65\%&0.87\% & 5.78\% & 0.58\%  \\
\hline
Ours-WD & \bf{2.76\%} & \bf{0.58\%} & \bf{5.98\%} & \bf{1.55\%} & \bf{5.86\%} & \bf{0.71\%} & \bf{6.40\%} & \bf{0.75\%} & \bf{3.85\%}& \bf{0.46\%}\\
\hline
\end{tabular}}
\end{center}
\end{table*}

Three performance indexes are adopted to evaluate the recognition methods:

False acceptance rate (FAR):
\begin{equation}
FAR = \frac{N_{FA}}{N_{NEG}}
\end{equation}
where $N_{FA}$ is the number of false accepted pairs, $N_{NEG}$ is the total number of negative pairs.

False rejection rate (FRR):
\begin{equation}
FRR = \frac{N_{FR}}{N_{POS}}
\end{equation}
where $N_{FR}$ is the number of false rejected pairs, $N_{POS}$ is the total number of positive pairs.

Equal error rate (EER):
EER is the working point on the DET curve where FAR equals to FRR.

For all iris images, the preprocessing procedure contains three steps: 1) eye detection by Haar-like Adaboost detectors~\cite{Viola2004Robust}, 2) iris boundary localization using the method in~\cite{Zhaofeng2009Toward}, and 3) iris normalization by rubber sheet model~\cite{Daugman1993High}. 
%Modified 4-th Minor: Circular
Circular iris images are normalized to a rectangle with $128 \times 256$ resolution.

Five databases are used for experiments: 

(1) ND CrossSensor Iris 2013 Dataset-LG4000 (ND-LG4000)~\cite{NDCS2013}. This database is one of the most popular iris databases for iris recognition research. It contains 29,986 iris samples from 1,352 classes. The training set of this database consists of images of the first 676 classes. The test set, after removing some falsely normalized images, consists of the remaining 676 classes and contains 114,243 genuine pairs and 57,554,187 imposter pairs.

(2) CASIA Iris Image Database V4-Distance (CASIA-Distance)~\cite{DataCASIAv4}. Images of this database are acquired from 3 meters away. It contains 2,446 iris samples from 284 classes. The training set of this database consists of images of the first 142 classes. The test set, after removing some falsely normalized images, consists of the remaining 142 classes and contains 12,617 genuine pairs and 1,695,859 imposter pairs.

(3) CASIA-Iris-M1-S2 (CASIA-M1-S2)~\cite{DataCASIAv4}. Images of this database are acquired by mobile devices at three different distances. It contains 6,000 iris samples from 400 classes. The training set of this database consists of images of the first 200 classes. The test set consists of the remaining 200 classes and contains 21,000 genuine pairs and 4,477,500 imposter pairs.

(4) CASIA-Iris-M1-S3 (CASIA-M1-S3)~\cite{DataCASIAv4}. Images of this database are acquired by mobile devices, different from CASIA-Iris-M1-S2. It contains 3,600 iris samples from 720 classes. The training set of this database consists of images of the first 360 classes. The test set consists of the remaining 360 classes and contains 3,600 genuine pairs and 1,615,500 imposter pairs.

(5) CASIA Iris Image Database V4-Lamp (CASIA-Lamp)~\cite{DataCASIAv4}. This database contains 16,212 iris samples from 819 classes. A lamp close to the subject was turned on/off to introduce elastic deformation of the iris due to pupil expansion and contraction under different illumination conditions. The training set of this database consists of the first 409 classes. The test set, after removing some falsely normalized images, consists of the remaining 410 classes and contains 76,387 genuine pairs and 32,700,296 imposter pairs.

%----------------------------------------------------------------------------------

\subsubsection{Comparisons to State-of-the-Art Approaches}

%Modified 4-th Minor: local area patterns &  large region patterns
\begin{figure}[h]
\begin{center}
\includegraphics[width=0.9\linewidth]{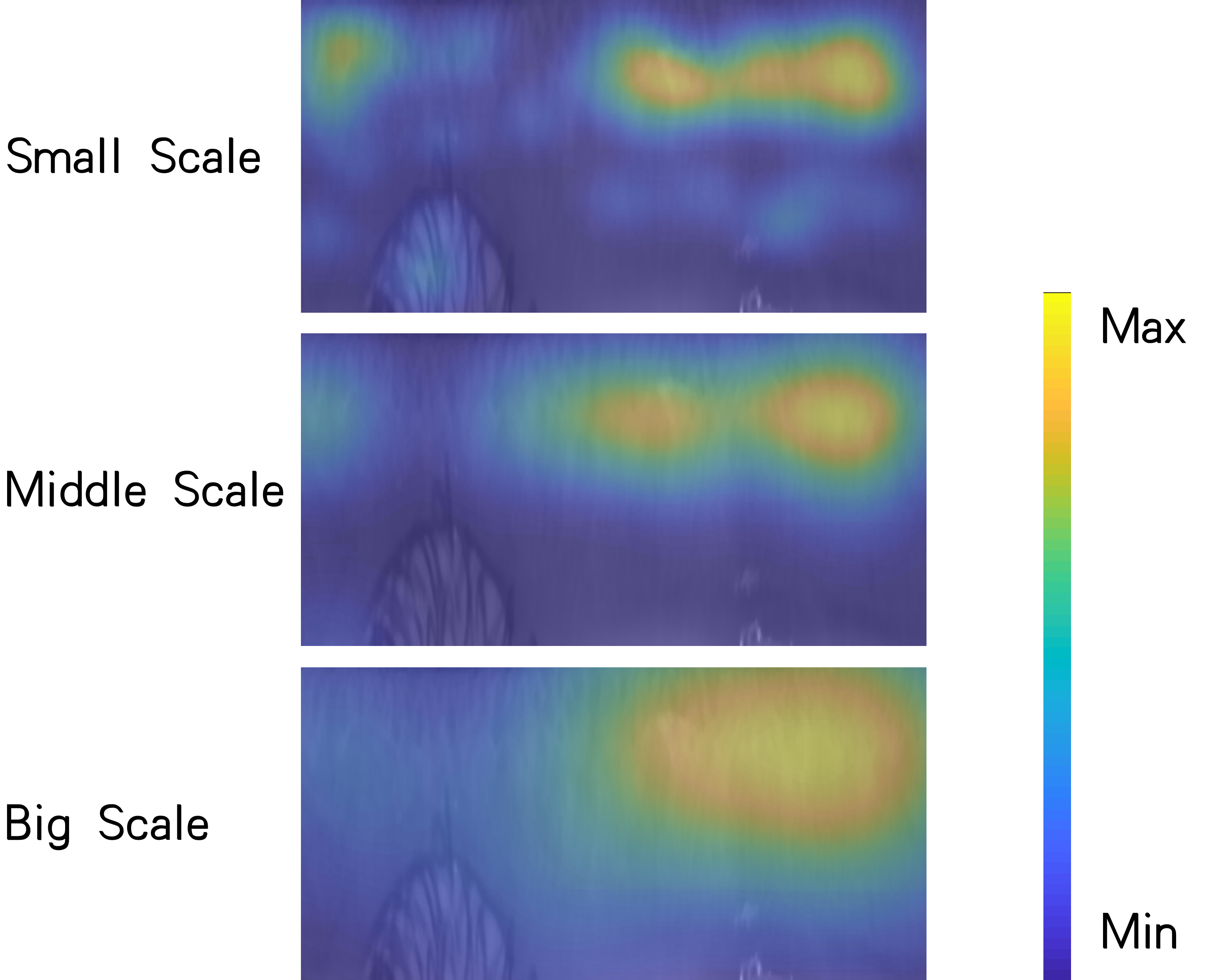}
\end{center}
   \caption{Visualization of the saliency maps of FGs of an iris sample from CASIA-Lamp. The saliency maps are computed by superposing the effective receptive field (ERF)~\cite{Wenjie2017Understanding} of all nodes. As we can observe from the saliency maps, the occluded areas caused by eyelids and eyelashes are avoided. The small-scale FG focuses on local area patterns, and the large-scale FG focuses on large region patterns.}
\label{fig:iris_graph}
\end{figure}

The proposed method is compared with four state-of-the-art approaches: log-Gabor~\cite{L2003}, ordinal measures (OMs)~\cite{Zhenan2009Ordinal}, UniNet~\cite{Zhao2017Towards} and MaxoutCNNs~\cite{Zhang2018Deep}. Handcrafted approaches and deep learning approaches are both taken into consideration.
Open source implementation of USIT\footnote{http://www.wavelab.at/sources/Rathgeb16a/} of log-Gabor~\cite{L2003} is adopted. 
We reproduced OMs in~\cite{Zhenan2009Ordinal}, and have tried our best to optimize the parameters for fair comparison.
Authoritative implementations of UniNet~\cite{Zhao2017Towards} and MaxoutCNNs~\cite{Zhang2018Deep} are used, but the parameters are retrained following the configurations of experiments of this paper, and we have tried our best to optimize the models.

Two test configurations are incorporated to evaluate the proposed method: within-database (WD) and cross-database (CD) configuration. In the within-database (WD) configuration, models trained on ND-LG4000 were utilized as the pretrained model, and the models were fine-tuned on the other databases. In the cross-database (CD) configuration, ND-LG4000 was utilized as training set, and the trained models were tested without any fine-tuning process.
The purpose of cross-database (CD) configuration is to evaluate the generalization capability of the proposed method.

The results are shown in Fig.~\ref{fig:roc_iris} and Tab.~\ref{tab:iris}. Significant improvements from the proposed framework can be found for all five databases.

The saliency map of FGs of an iris sample from CASIA-Lamp is visualized in Fig.~\ref{fig:iris_graph}. 
%Nodes generated by Graph Block 1 are shown in the first row, nodes generated by Graph Block 2 in the second row and nodes generated by Graph Block 3 in the third row. 
Occluded regions caused by eyelids and eyelashes, which are useless for iris recognition, are suppressed. 
%Modified 4-th Minor: local area patterns &  large region patterns
The small-scale FG focuses on local area patterns, and the large-scale FG focuses on large region patterns.

%----------------------------------------------------------------------------------

\subsubsection{Occluded Iris Recognition}
\label{exp:occ_iris}
\begin{figure}[h]
\begin{center}
\includegraphics[width=1.\linewidth]{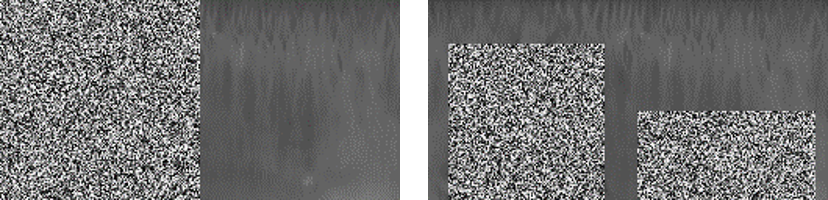}
\end{center}
   \caption{Two kinds of occlusions are used in the experiment of occluded iris recognition. The covered areas of this experiment are set according to the locations of eyelids and eyelashes, which are the main culprits of occlusions, in normalized iris images.}
\label{fig:iris_occ}
\end{figure}

\begin{figure}[h]
\begin{center}
\includegraphics[width=1.\linewidth]{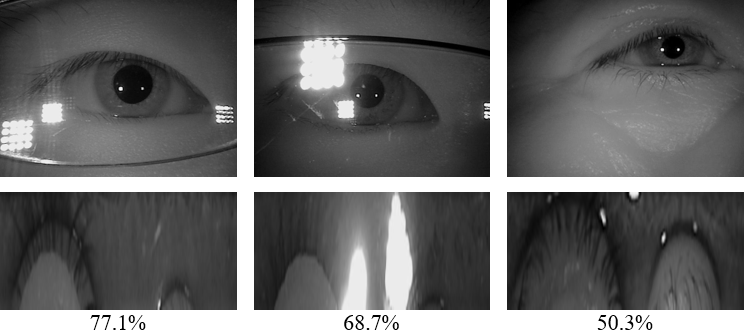}
\end{center}
   \caption{Occluded iris samples in CASIA Iris Image Database V4-Thousand. The first row is the raw eye images. The second row is the normalized iris images. Their USABLE\_IRIS\_AREA are shown at the bottom.}
\label{fig:iris_occ_real}
\end{figure}

The occluded iris recognition experiment is launched on two databases. The first is the test database of ND CrossSensor Iris 2013 Dataset-LG4000. Special areas of iris samples are covered by random noise to simulate the occluded situations on this database, as shown in Fig.~\ref{fig:iris_occ}.
The second is CASIA Iris Image Database V4-Thousand~\cite{DataCASIAv4}. This database contains 20,000 iris samples from 2,000 classes. Various real occlusions, including eyelid, eyelash, and light spot, are contained in this database, as shown in Fig.~\ref{fig:iris_occ_real}.

%The occluded iris recognition experiment is launched on the test database of ND CrossSensor Iris 2013 Dataset-LG4000. Special areas of iris samples are covered by random noise to simulate the occluded situations during this experiment, as shown in Fig.~\ref{fig:iris_occ}.
For the test database of  ND CrossSensor Iris 2013 Dataset-LG4000, the replaced areas are set according to the locations of eyelids and eyelashes in normalized iris images.
Two kinds of occlusion situations shown in Fig.~\ref{fig:iris_occ} are randomly selected with the same probability for a special iris sample, while the percentage of covered area remains the same. 
Masks, which contain the prior knowledge of the occluded region, are unavailable in these experiments for fair comparison.
Comparisons to mask strategy are presented in Section~\ref{exp:mask}.

% Modified 4-th Minor: (2)...
Two protocols are adopted in the experiments on ND CrossSensor Iris 2013 Dataset-LG4000: (1) the occluded areas are cropped before feature extraction (the pixels of occluded areas are set to 0 if the input shape of the recognition model is constant), and (2) the occluded areas are also fed to the feature extractor.
Note that none of the models used in this experiment are trained on the occluded database.

The results of two test protocols under $30\%$ occluded areas are shown in Tab.~\ref{tab:roc_occ}. The performance gap of the proposed framework between protocols (1) and (2) is much smaller than that for the other methods. 
For handcrafted approaches, the performance of protocol (2) is degraded because features extracted from occluded regions are interfused. For the deep learning-based methods, occluded regions of protocol (2) significantly increase the distance between intraclass samples, and the generalization abilities of these models in occluded situations are quite limited.
%The proposed framework works well because the nodes corresponding to occluded regions are adaptively removed.
%The results indicate that the proposed framework offers desirable generalization ability in occluded situations.

%\begin{figure}[h]
%\begin{center}
%\includegraphics[width=\linewidth]{iris_occ_plus.png}
%\end{center}
 %  \caption{DET curves under occluded situations. The percentage of the occluded area is $30\%$. The results show that the gap between protocols (1) and (2) of the proposed method is much smaller than that for the other methods.}
%\label{fig:roc_occ}
%\end{figure}

%\begin{figure}[h]
%\begin{center}
%\includegraphics[width=\linewidth]{roc_occ33.png}
%\end{center}
%  \caption{DET curves under occluded situations. The occluded areas are remained. The percentage of occluded area is $30\%$.}
%\label{fig:roc_occ}
%\end{figure}

\begin{table}
\begin{center}
\caption{FRR and EER of simulative occluded situations. The percentage of the occluded area is $30\%$.}
\label{tab:roc_occ}
\setlength{\tabcolsep}{5mm}
{
\begin{tabular}{c|c|c}
\hline
& \bf{FRR@FAR=0.1\%} &\bf{EER} \\
\hline
log-Gabor (1) \cite{L2003}& 26.73\% & 7.05\%  \\
%\hline
log-Gabor (2) \cite{L2003}& 41.76\% & 27.25\%  \\
\hline
OMs (1) \cite{Zhenan2009Ordinal}& 16.72\% &5.3\%  \\
%\hline
OMs (2) \cite{Zhenan2009Ordinal}& 38.67\% &26.29\%  \\
\hline
UniNet (1) \cite{Zhao2017Towards}& 5.81\% & 3.98\%  \\
%\hline
UniNet (2) \cite{Zhao2017Towards}& 45.33\% & 8.32\%  \\
\hline
MaxoutCNNs (1) \cite{Zhang2018Deep}& 76.25\% & 9.61\%  \\
%\hline
MaxoutCNNs (2) \cite{Zhang2018Deep}& 73.68\% & 21.8\%  \\
\hline
Ours (1)& 2.28\% & 1.03\%  \\
%\hline
Ours (2)& \bf{1.48\%} & \bf{0.80\%}  \\
\hline
\end{tabular}}
\end{center}
\end{table}

\begin{figure}[h]
\begin{center}
\includegraphics[width=\linewidth]{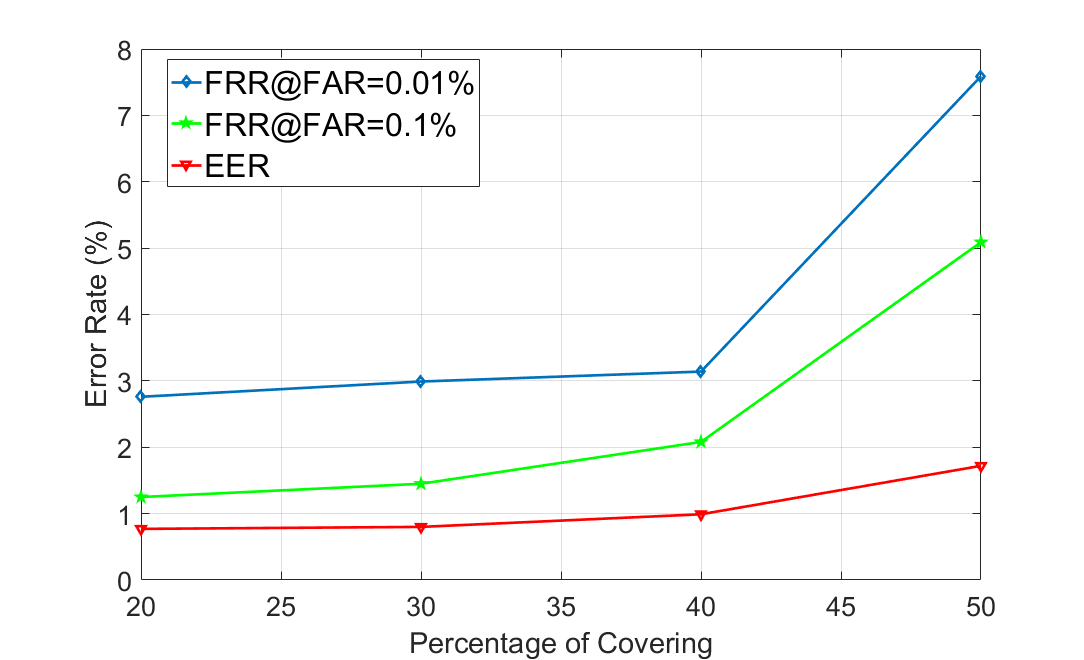}
\end{center}
   \caption{Performance of the proposed method under different percentages of occlusion.}
\label{fig:occ}
\end{figure}

An additional experiment is conducted to evaluate the influence of the occluded areas on the proposed framework. 
The performance of the proposed framework under different occluded sizes is provided in Fig.~\ref{fig:occ}. It is not surprising to observe that the performance of the proposed framework degrades when the occlusion size grows.
% Modified 4-th Minor: the occluded areas are over 40\%
The performance declines dramatically only when the occluded areas are over 40\%.% areas are occluded. 
One possible reason for this is that the threshold value of dynamic matching is set by the mean of similarities of nodes, as described in Section~\ref{med:dgm}, which means that the expected percentage of node removal is approximately 50\%. 
The proposed framework can function well when the percentage of the occlusion area is lower than 50\%, but features of occluded areas may be introduced when the percentage of the occluded area is higher than 50\%.

This phenomenon indicates that the strategy of dynamic matching is not optimal when the occlusion is extremely serious, which is a limitation of the proposed framework. %And this straightforward strategy can be improved by more robust methods in later research.

\begin{table*}
\begin{center}
\caption{FRR and EER of iris recognition with real occlusions.}
\label{tab:roc_occ_real}
\setlength{\tabcolsep}{5mm}
{
\begin{tabular}{c|c|c|c|c|c|c}
\hline
& \multicolumn{3}{c|}{\bf{FRR@FAR=0.1\%}} & \multicolumn{3}{c}{\bf{EER}} \\
\hline
\bf{USABLE\_IRIS\_AREA} &\bf{70\%-80\%} &\bf{60\%-70\%}&\bf{50\%-60\%}&\bf{70\%-80\%}&\bf{60\%-70\%}&\bf{50\%-60\%} \\
\hline
log-Gabor  \cite{L2003}& 18.60\%& 27.26\% & 27.14\% & 4.85\%& 7.03\%  & 7.27\%\\
\hline
OMs  \cite{Zhenan2009Ordinal}& 17.65\%& 23.99\%& 24.86\% &4.71\%&6.49\%&6.87\%  \\
\hline
UniNet  \cite{Zhao2017Towards}& 10.24\%& 22.69\% & 24.40\% &2.65\%& 4.32\% & 6.38\%   \\
\hline
MaxoutCNNs \cite{Zhang2018Deep}& 86.81\%& 86.91\% & 89.65\%& 14.70\%& 15.22\%&15.73\%  \\
\hline
Ours & \bf{2.21\%} & \bf{3.75\%} & \bf{5.25\%}&\bf{0.65\%} & \bf{1.15\%} & \bf{1.99\%}\\
\hline
\end{tabular}}
\end{center}
\end{table*}

To test the performance of the proposed framework on iris images with real occlusions, the experiments are conducted on CASIA Iris Image Database V4-Thousand.
The samples in this database are divided  into subsets according to their USABLE\_IRIS\_AREA defined by ISO/IEC 29794-6~\cite{IrisISO}:

\begin{equation}
(1-\frac{N_{occluded}}{N_{iris}}) \times 100\%
\end{equation}

where $N_{iris}$ is the area of iris, $N_{occluded}$ is the area of occluded iris.
In this experiment, $N_{iris}$ is calculated according to the inner and outer boundary of iris, $N_{occluded}$ is calculated according to the ground truth iris mask. 
The USABLE\_IRIS\_AREA is the fraction of the iris portion of the image that is not occluded.
Three subsets are adopted for testing, the range of USABLE\_IRIS\_AREA are 70\% - 80\%, 60\% - 70\%, and 50\% - 60\% respectively.
To evaluate the generalization ability of the proposed method, the models in this experiment are not re-trained on CASIA Iris Image Database V4-Thousand.

The results are shown in Tab.~\ref{tab:roc_occ_real}.
The proposed framework significantly outperforms the compared methods in all three subsets.
The proposed framework works well in both simulated and real occlusion scenarios because the nodes corresponding to occluded regions are adaptively removed.
The results indicate that the proposed framework offers desirable generalization ability in real-world occluded situations.

%-----------------------------------------------------------------------------------------------
%-----------------------------------------------------------------------------------------------

\subsection{Face Recognition Experiments}

\begin{table*}[h]
\begin{center}
\caption{Results of face recognition.}
\label{tab:face}
{
\begin{tabular}{c|c|c|c|c}
\hline
& TAR@FAR=1\%& TAR@FAR=0.1\% & TAR@FAR=0&100\%-EER \\
\hline
 \multicolumn{5}{c}{\bf{LFW}}\\
\hline
ArcFace (ResNet50)~\cite{Deng2018ArcFace}& 99.64\% & 99.33\% & 97.23\% & 99.56\%  \\
\hline
ArcFace (ResNet101)~\cite{Deng2018ArcFace}& 99.75\% & 99.54\% & 97.90\% & 99.67\%  \\
\hline
\hline
Ours (ResNet50) & 99.67\% & 99.20\% & 98.15\% & 99.56\%  \\
\hline
Ours (ResNet101) & \bf{99.87\%} &\bf{99.71\%} & \bf{98.22\%} & \bf{99.77\%}  \\
\hline

 \multicolumn{5}{c}{\bf{MegaFace}}\\
\hline
ArcFace (ResNet50)~\cite{Deng2018ArcFace}& 96.92\% & 94.23\% & 89.36\% & 87.64\%  \\
\hline
ArcFace (ResNet101)~\cite{Deng2018ArcFace}& \bf{98.61\%} & 96.22\% & 78.18\% & 75.16\%  \\
\hline
\hline
Ours (ResNet50) & 97.26\% & 95.15\% & \bf{93.91\%} & \bf{90.62\%}  \\
\hline
Ours (ResNet101) & 98.47\% &\bf{96.69\%} & 89.80\% & 83.99\%  \\
\hline
\end{tabular}}
\end{center}
\end{table*}

\begin{figure*}[h]
\begin{center}
\includegraphics[width=\linewidth]{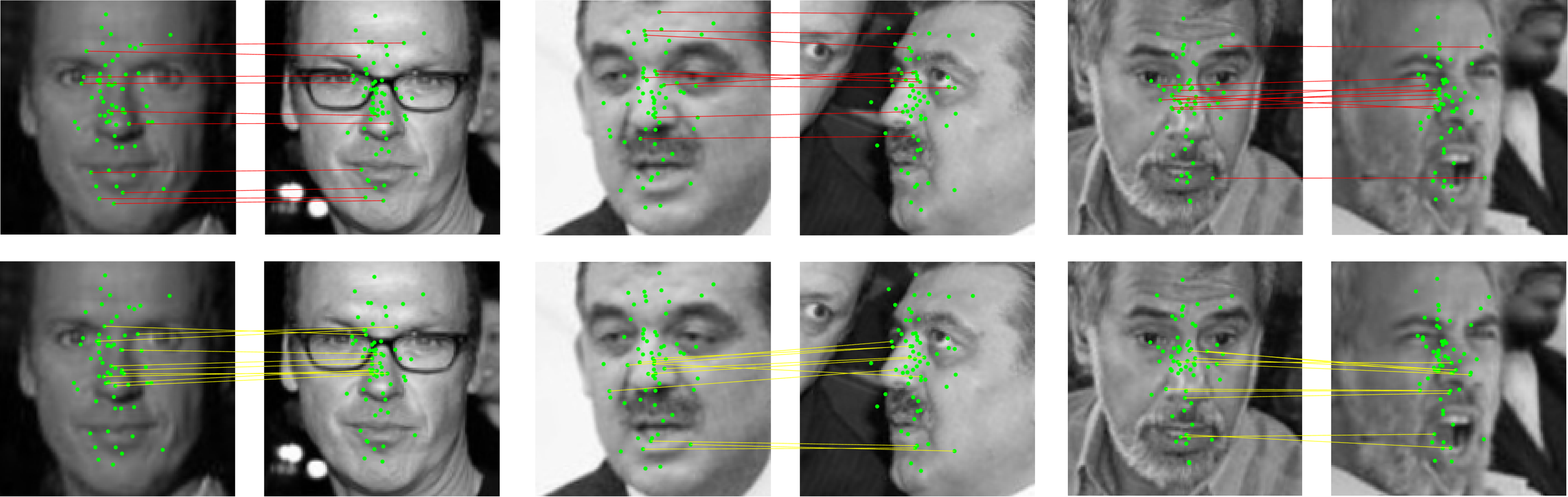}
\end{center}
\setlength{\abovecaptionskip}{0pt}
\setlength{\belowcaptionskip}{0pt}
   \caption{Matched node pairs of face samples (small scale). The pairs of nodes linked by red lines are the 10 pairs with the highest similarity scores. The pairs of nodes linked by yellow lines are the 10 pairs with the lowest similarity scores. The 10 node pairs with highest/lowest similarity scores show the regions deemed as most similar/dissimilar by the framework.
These examples show that FGs enable a more illustrative and reasonable inference by showing local patch similarities with representations (best viewed in color).}
\label{fig:face_match}
\end{figure*}

\subsubsection{Comparisons to State-of-the-Art Methods}
The proposed method is compared with ArcFace~\cite{Deng2018ArcFace}, which is a state-of-the-art method for face recognition. ResNet50 and ResNet101 are utilized as backbone. 
MS-DGR is integrated into these models.
For ResNet50, the first convolutional block contains 14 convolutional layers, the second consists of 28 convolutional layers, and the third consists of the 6 remaining convolutional layers. For ResNet101, the first convolutional block contains 32 convolutional layers, the second consists of 60 convolutional layers, and the third consists of the 6 remaining convolutional layers.
%Modified 4-th Minor: when constructing FGs from respective feature maps
The node numbers are set to 64, 32 and 16 when constructing FGs from respective feature maps. 
Stochastic gradient descent with momentum is adopted for optimization. Training is started with a learning rate of 0.01 and divided by 10 every 10 epoch and stopped at the 30th epoch. The size of the mini-batch is 128. The weight decay is 0.0001.

The data preprocessing method of ArcFace~\cite{Deng2018ArcFace} is used, and the training databases\footnote{https://github.com/deepinsight/insightface} of ArcFace~\cite{Deng2018ArcFace} are adopted for fair comparison.
Labeled Faces in the Wild~\cite{Learned2014Labeled} (LFW) and MegaFace~\cite{Kemelmacher2016The} are selected as the test sets.

The results are shown in Tab.~\ref{tab:face}. The proposed method is better than ArcFace~\cite{Deng2018ArcFace} in most cases. Along with the decreasing FAR, the advantage of our method becomes more salient, which indicates the effectiveness of the proposed graph module in our framework for face recognition.

Small-scale FGs of example face images are visualized in Fig.~\ref{fig:face_match}. The first row shows the 10 pairs of nodes with the highest similarity scores. The second row shows the 10 node pairs with the lowest similarity scores. 
These examples show the regions that are deemed most similar/dissimilar by the framework, which is the underlying reason for recognition decisions.
%All the 10 pairs nodes with highest similarity scores lie in the left area of the right eye which indicates that the network ``thinks'' this is the most similar region of the two images. 
%There are two significant differences between the two images: the left area of the nose is occluded in the right image because of the pose, and mouths in two images are different. Most of the 10 node pairs with lowest similarity scores lie in these two areas which means the network ``finds'' these two differences. 
These examples show that FGs produce more illustrative and reasonable inferences by showing local patch similarities with representations.
For example, in the left pair of images, the most similar nodes pairs are uniformly distributed over the whole face since the poses and expressions are similar.
The most dissimilar nodes scatter around the eyes because of the glasses.
%More examples are shown in supplementary material.

% Modified 4-th Minor: detects
Meanwhile, these examples show that the SLN, which detects the spatial locations of nodes, is robust to illumination (the left pair of Fig.~\ref{fig:face_match}), head pose (the middle pair of Fig.~\ref{fig:face_match}), and expression (the right pair of Fig.~\ref{fig:face_match}).
SLN chooses the discriminative regions of the image pairs in these scenarios.
For example, in the middle pair of images, nodes of the right image are concentrated on the right-half face because the left-half face is not visible.

%-----------------------------------------------------------------------------------------------

\subsubsection{Occluded Face Verification}
\label{sec:facever}

%Receiver operating characteristic (ROC) curves are adopted for occluded face verification to evaluate the proposed framework.
Light CNN (LCNN)~\cite{Xiang2018A} is adopted to build the architecture for occluded face recognition. The LCNN-9 version that contains 9 convolutional layers is selected. 

Similar to ArcFace experiments, the LCNN model is split into three convolutional blocks for the integration of graph modules. 
The first convolutional block contains 5 convolutional layers, the second consists of 2 convolutional layers, and the third consists of the remaining 2 convolutional layers.
The node numbers of the three FGs are set as 128, 64, and 64.
%Original Light CNN-9 is selected as the baseline method for comparison naturally.
Stochastic gradient descent with momentum is adopted for optimization. Training is started with a learning rate of 0.001 and divided by 2 every 10 epoch and stopped at the 40th epoch. The size of the mini-batch is 128. The weight decay is 0.0001.

The data preprocessing method of LCNN is used, and the training set is CASIA-WebFace~\cite{Yi2014Deep}.
%CASIA-WebFace is a public database which contains 494,414 face images from 10,575 subjects.
Horizontal mirror operation is conducted for data augmentation because faces have a nearly symmetric structure.

\begin{figure}[h]
\begin{center}
\includegraphics[width=\linewidth]{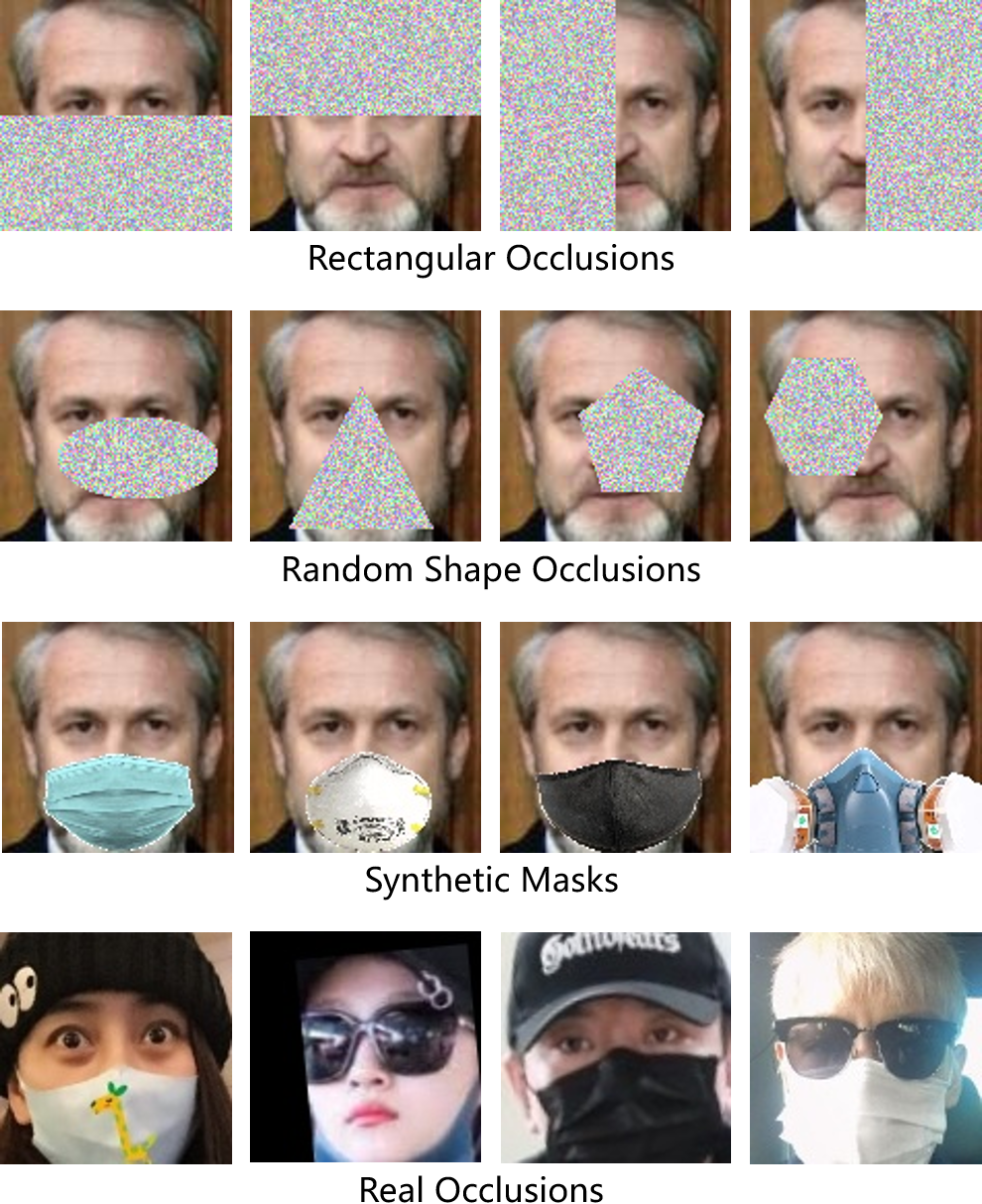}
\end{center}
   \caption{Three kinds of simulated occlusions and real occlusion used in the occluded face samples experiment.}
\label{fig:face_occ}
\end{figure}

Both simulated and real occlusions are taken into consideration.
For simulated occlusions, a simulated occluded face database named Occluded-LFW, which is based on the Labeled Faces in the Wild~\cite{Learned2014Labeled} (LFW) database, is used for evaluation. LFW contains 13,233 images from 7,749 individuals. Face images of LFW vary in terms of pose, expression, and illumination. 
After the same preprocessing of the training images, three kinds of simulated occlusions are taken into consideration, as shown in the first three rows of Fig.~\ref{fig:face_occ}.
In the first case, rectangular areas of face images are covered by random noises to simulate the occluded situations during this experiment.
Four regions in the first row of Fig.~\ref{fig:face_occ}, right, left, upper, and bottom, are randomly covered with the same probability during this experiment.
In the second case, face images are covered by random shapes of noises.
Four shapes shown in the second row of Fig.~\ref{fig:face_occ} are randomly selected with the same probability during this experiment. 
The location of the noises is also randomly selected.
In the third case, four kinds of masks, the surgical mask, N95, the cloth mask, and the gas mask, are overlaid on the face images according to face landmarks. 
The four kinds of masks are randomly selected with the same probability during this experiment. 
We follow the Labeled Faces in the Wild (LFW) benchmark protocol\footnote{http://vis-www.cs.umass.edu/lfw/pairs.txt}, where 3,000 positive pairs and 3,000 negative pairs of images are selected for face verification. 
For each pair, one image is from the Occluded-LFW database, and the other is from the LFW database.

For real occlusions, Real-World Masked Face Dataset (RMFD)~\cite{Zhongyuan2020Masked} is adopted as the test database.
The real occlusions of RMFD contain masks, hats, and sunglasses, as shown in the last row of Fig.~\ref{fig:face_occ}.
Some of the samples are quite difficult.
The subset of RMFD for verification contains 3,589 positive pairs and 3,589 negative pairs. One image of each pair is occluded and the other is not.
To evaluate the generalization ability of the proposed method, the models in this experiment are not re-trained on the occluded database.

The proposed framework is compared with two occluded face recognition methods: I2C \cite{Hu2013Robust} and DFM \cite{He2018Dynamic}. LCNN are also evaluated as a baseline model. 
%Experiments with two different percentages of occluded area are launched: 30\% and 50\%. 
The results of occluded face recognition are shown in Tab.~\ref{tab:face_ver}.
The proposed framework outperforms the compared methods under all four kinds of occlusions.

%\begin{figure}[h]
%\begin{center}
%\includegraphics[width=\linewidth]{face_ver.png}
%\end{center}
%   \caption{ROC curves of occluded face verification: (a) the percentage of occluded area is 30\%; (b) the %percentage of occluded area is 50\%.}
%\label{fig:face_ver}
%\end{figure}

\begin{table}
\begin{center}
\caption{Performance of occluded face verification.}
\label{tab:face_ver}
\setlength{\tabcolsep}{5mm}
{
\begin{tabular}{c|c|c}
\hline
& TAR@FAR=0.1\% &100\%-EER \\
\hline
\hline
 \multicolumn{3}{c}{\bf{Rectangle, Occluded Area: 30\%}}\\
\hline
LCNN-9 \cite{Xiang2018A}& 22.47\% & 85.97\%  \\
\hline
I2C \cite{Hu2013Robust}& 0.4\% & 64.3\%  \\
\hline
DFM \cite{He2018Dynamic}& 61.21\% & 90.03\%  \\
\hline
Ours& \bf{73.14\%} &\bf{93.83\%}  \\
\hline

 \multicolumn{3}{c}{\bf{Rectangle, Occluded Area: 50\%}}\\
\hline
LCNN-9 \cite{Xiang2018A}& 8.47\% & 76.47\%  \\
\hline
I2C \cite{Hu2013Robust}& 0.3\% & 63.19\%  \\
\hline
DFM \cite{He2018Dynamic}& 12.8\% & 84.59\%  \\
\hline
Ours& \bf{30.03\%} &\bf{84.77\%}  \\
\hline
\hline
 \multicolumn{3}{c}{\bf{Random Shape, Occluded Area: 20\%}}\\
\hline
LCNN-9 \cite{Xiang2018A}& 9.3\% & 78.1\%  \\
\hline
DFM \cite{He2018Dynamic}& 25.3\% & 85.47\%  \\
\hline
Ours& \bf{37.57\%} &\bf{86.4\%}  \\
\hline
\multicolumn{3}{c}{\bf{Random Shape, Occluded Area: 30\%}}\\
\hline
LCNN-9 \cite{Xiang2018A}& 1.27\% & 61.1\%  \\
\hline
DFM \cite{He2018Dynamic}& 9.87\% & 76.13\%  \\
\hline
Ours& \bf{35.53\%} &\bf{77.63\%}  \\
\hline
\hline
 \multicolumn{3}{c}{\bf{Synthetic Mask}}\\
\hline
LCNN-9 \cite{Xiang2018A}& 15.41\% & 81.04\%  \\
\hline
I2C \cite{Hu2013Robust}& 0.91\% & 67.38\%  \\
\hline
DFM \cite{He2018Dynamic}& 31.36\% & 86.96\%  \\
\hline
Ours& \bf{40.49\%} &\bf{88.85\%}  \\
\hline
\hline
 \multicolumn{3}{c}{\bf{Real Occlusions}}\\
\hline
LCNN-9 \cite{Xiang2018A}& 0.48\% & 60.28\%  \\
\hline
DFM \cite{He2018Dynamic}& 1.53\% & 67.56\%  \\
\hline
Ours& \bf{33.87\%} &\bf{83.64\%}  \\
\hline

\end{tabular}}
\end{center}
\end{table}

%-----------------------------------------------------------------------------------------------

\subsubsection{Occluded Face Identification}

The same models of occluded face verification are adopted for occluded face identification. 
For simulated occlusions, the probe set consists of 1,000 face images from 1,000 identifications, and the gallery set consists of 1,000 face images with shared identities. The images of the probe set are from the Occluded-LFW database, and the images of the gallery set are from the LFW database.
For real occlusions, the subset of RMFD for recognition contains 5,000 occluded samples from 525 classes as the probe set and 90,000 non-occluded samples from the same 525 classes as the gallery set.

Experiments of occluded face identification are also launched under the three kinds of simulated occlusions and the real occlusions.
The results are shown in Tab.~\ref{tab:face_ide}.
Similar to the experiments of face verification, obvious promotions can be observed in most settings between the compared methods and the proposed framework, which demonstrate the effectiveness of the proposed framework.

\begin{table}
\begin{center}
\caption{Performance of occluded face identification.}
\label{tab:face_ide}
\setlength{\tabcolsep}{5mm}
{
\begin{tabular}{c|c|c|c}
\hline
& Rank-1 &Rank-5 &Rank-10 \\
\hline
\hline
 \multicolumn{4}{c}{\bf{Rectangle, Occluded Area: 30\%}}\\
\hline
LCNN-9 \cite{Xiang2018A}& 65.5\% & 75.9\% & 78.6\%  \\
\hline
I2C \cite{Hu2013Robust}& 8.4\% & 13.1\% & 15.5\%  \\
\hline
DFM \cite{He2018Dynamic}& 74.7\% & 80.2\% & 84.4\%  \\
\hline
Ours& \bf{80.2\%} &\bf{87.3\%} & \bf{89.5\%}  \\
\hline

 \multicolumn{4}{c}{\bf{Rectangle, Occluded Area: 50\%}}\\
\hline
LCNN-9 \cite{Xiang2018A}& 26.2\% & 40.5\% & 47.3\%  \\
\hline
I2C \cite{Hu2013Robust}& 7.0\% & 10.7\% & 13.2\%  \\
\hline
DFM \cite{He2018Dynamic}& 30.2\% & 47.3\% & 52.9\%  \\
\hline
Ours& \bf{37.2\%} &\bf{54.0\%} & \bf{59.4\%}  \\
\hline
\hline
 \multicolumn{4}{c}{\bf{Random Shape, Occluded Area: 20\%}}\\
\hline
LCNN-9 \cite{Xiang2018A}& 36.9\% & 47.1\% & 51.9\%  \\
\hline
DFM \cite{He2018Dynamic}& 45.9\% & 62.1\% & 67.2\%  \\
\hline
Ours& \bf{46.8\%} &\bf{65.1\%} & \bf{67.8\%}  \\
\hline

 \multicolumn{4}{c}{\bf{Random Shape, Occluded Area: 30\%}}\\
\hline
LCNN-9 \cite{Xiang2018A}& 18.9\% & 36.8\% & 40.4\%  \\
\hline
DFM \cite{He2018Dynamic}& \bf{28.2\%} & 40.5\% & 46.2\%  \\
\hline
Ours& 27.6 &\bf{40.7\%} & \bf{46.3\%}  \\
\hline
\hline
 \multicolumn{4}{c}{\bf{Synthetic Mask}}\\
\hline
LCNN-9 \cite{Xiang2018A}& 55.5\% & 68.4\% & 73.9\%  \\
\hline
I2C \cite{Hu2013Robust}& 8.8\% & 13.3\% & 15.1\%  \\
\hline
DFM \cite{He2018Dynamic}& 68.2\% & 78.6\% & 82.1\%  \\
\hline
Ours& \bf{70.1\%} &\bf{79.3\%} & \bf{84.5\%}  \\
\hline
\hline
 \multicolumn{4}{c}{\bf{Real Occlusions}}\\
\hline
LCNN-9 \cite{Xiang2018A}& 6.1\% & 15.3\% & 21.8\%  \\
\hline
DFM \cite{He2018Dynamic}& 37.1\% & 38.4\% & 45.1\%  \\
\hline
Ours& \bf{40.7\%} &\bf{55.3\%} & \bf{61.1\%}  \\
\hline

\end{tabular}}
\end{center}
\end{table}

%-----------------------------------------------------------------------------------------------

\subsubsection{The Influence of Occlusion Size}

\begin{figure}[h]
\begin{center}
\includegraphics[width=\linewidth]{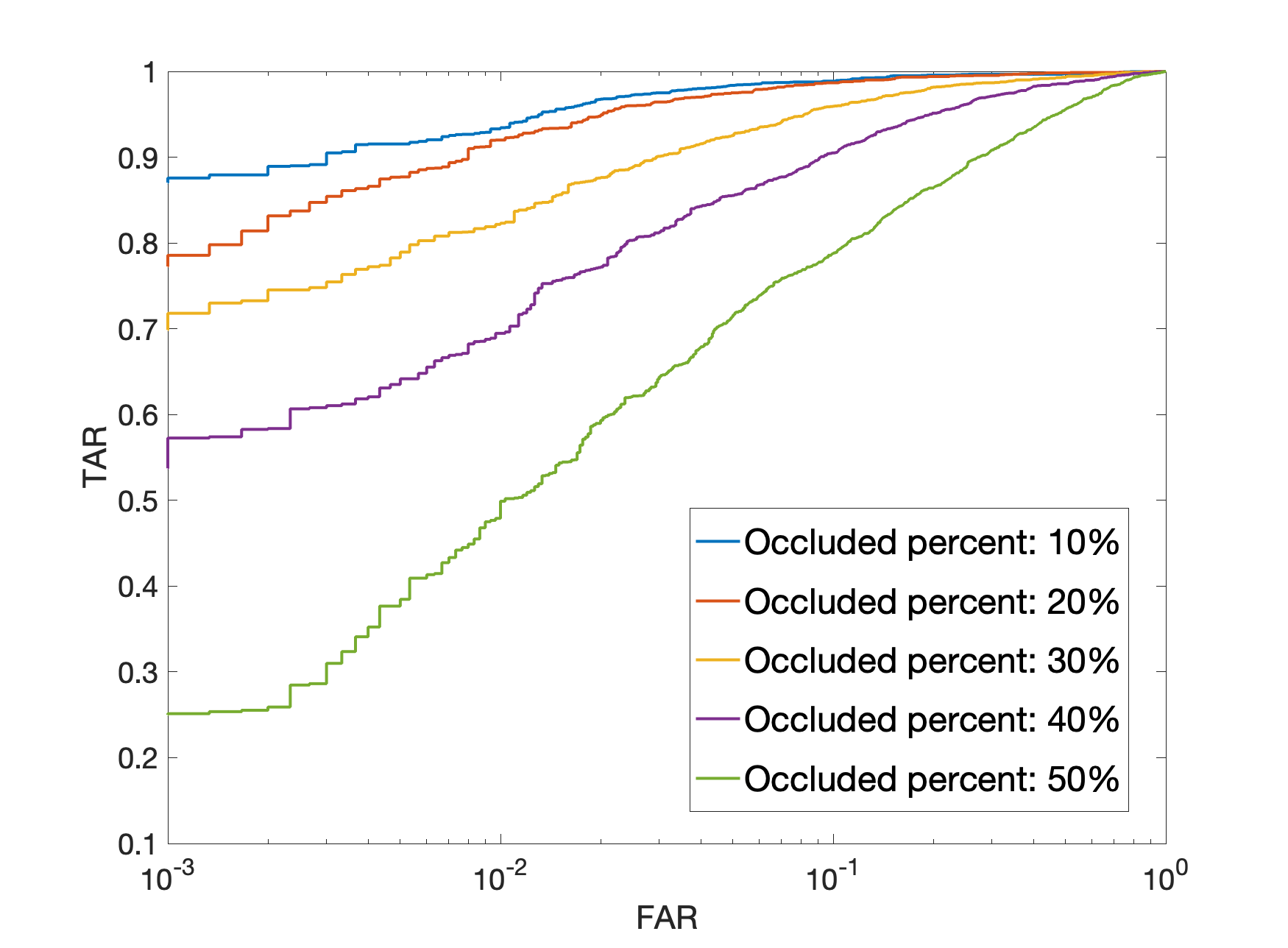}
\end{center}
   \caption{Performance of different percentages of occluded area.}
\label{fig:face_percent}
\end{figure}

To explore the influence of different sizes of occlusion, additional experiments are launched under different percentages of occluded area ranging from $10\%$ to $50\%$. 
Only the rectangular occlusions are adopted for the experiments on the size of occlusion.
The experiment protocols are the same as Section~\ref{sec:facever}.
The results are shown in Fig.~\ref{fig:face_percent} and Tab.~\ref{tab:face_percent}. 
The performance of the proposed method degrades as the size of occlusion grows because contextual information retained for recognition is gradually vanishing.

\begin{table}
\begin{center}
\caption{TAR and 100\%-EER of different percentages of occluded area.}
\label{tab:face_percent}
\setlength{\tabcolsep}{3mm}
{
\begin{tabular}{c|c|c}
\hline
Percentage of Occluded Area& TAR@FAR=0.1\% &100\%-EER \\
\hline
10\%& 87.57\% & 97.33\%  \\
\hline
20\%& 78.57\% &96.70\%  \\
\hline
30\%& 73.14\% & 93.83\%  \\
\hline
40\%& 57.27\% & 90.40\%  \\
\hline
50\%& 30.03\% & 84.77\%  \\
\hline
\end{tabular}}
\end{center}
\end{table}

%-----------------------------------------------------------------------------------------------

\subsubsection{The Influence of Occluded Position}

To explore the influence of different parts of the face on recognition, experiments on different occluded face parts are launched under the face verification protocol. 
Four regions shown in the first row of Fig.~\ref{fig:face_occ} are separately adopted to evaluate the performance of different parts of the face. The percentage of occluded area is fixed at 30\%.

\begin{figure}[h]
\begin{center}
\includegraphics[width=\linewidth]{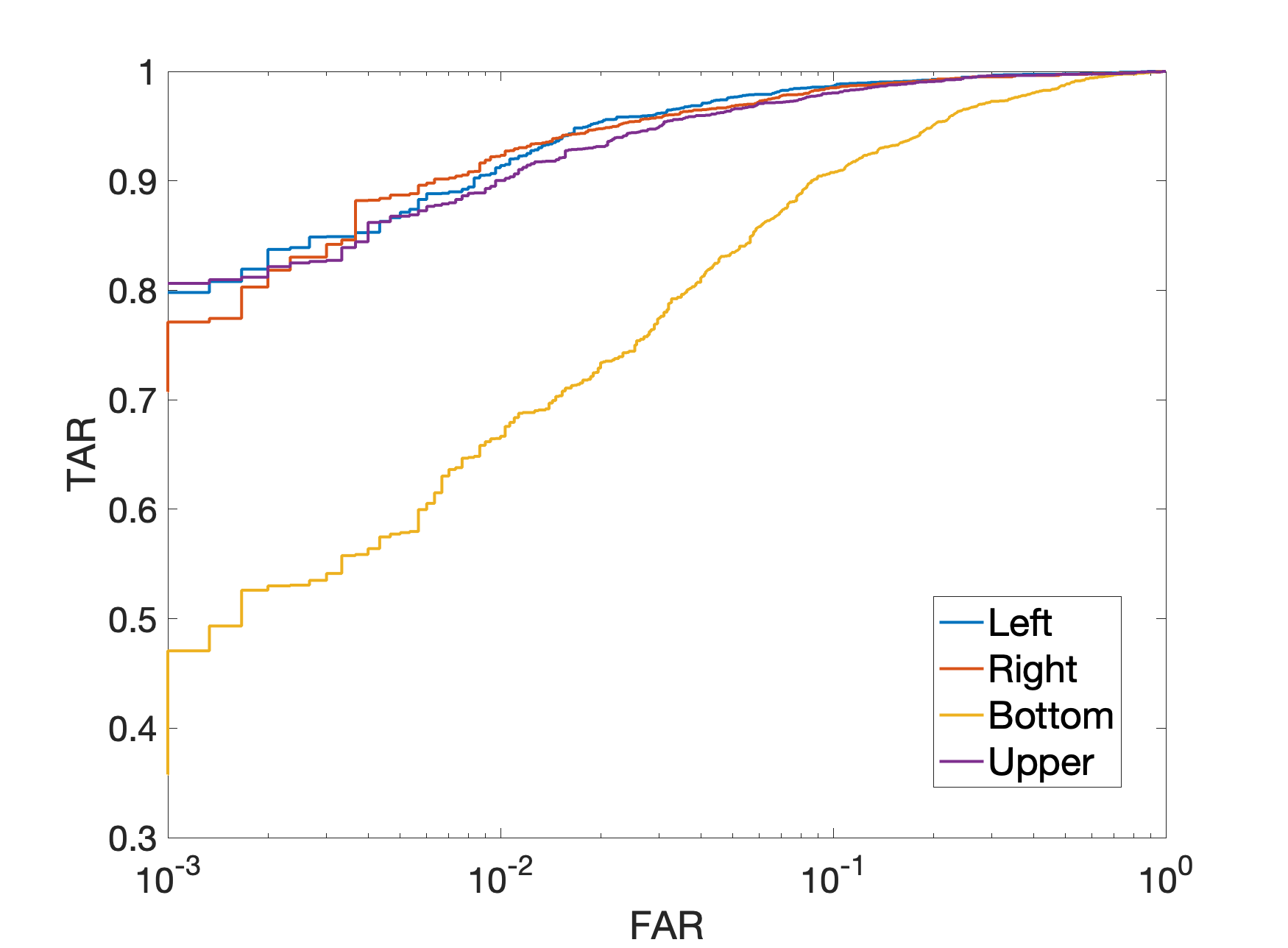}
\end{center}
   \caption{Performance of different parts of the face when the percentage of occluded area is 30\%.}
\label{fig:face_part}
\end{figure}

The results are shown in Fig.~\ref{fig:face_part} and Tab.~\ref{tab:face_part}. 
% Modified 4-th Minor: It can be observed that the performance is the worst when the upper part is occluded
It can be observed that the performance is the worst when the upper part is occluded, which indicates that the upper part of the face is more important for face recognition.
% Modified 4-th Minor: the performance of the others is nearly equal & for identity recognition
Another interesting observation is that the performance of the others is nearly equal, which verifies that facial images contain redundant information to some extent for identity recognition.
%part of face is close. The reason may be that faces have nearly horizontal symmetric structure. It is verified to some extent that facial image contains redundant information.

\begin{table}
\begin{center}
\caption{TAR and 100\%-EER of different parts of the face when the percentage of occluded area is 30\%.}
\label{tab:face_part}
\setlength{\tabcolsep}{4mm}
{
\begin{tabular}{c|c|c}
\hline
Visible Part of Face& TAR@FAR=0.1\% &100\%-EER \\
\hline
Left& 79.80\% & 96.57\%  \\
\hline
Right& 77.10\% &96.33\%  \\
\hline
Bottom& 47.07\% & 90.53\%  \\
\hline
Upper& 80.63\% & 95.97\%  \\
\hline
\end{tabular}}
\end{center}
\end{table}

%-----------------------------------------------------------------------------------------------
%-----------------------------------------------------------------------------------------------
\subsection{Comparisons to Masking Strategy}
\label{exp:mask}

In this section, comparative experiments to masking strategy are launched for comparison. The handcrafted approach and deep learning framework are both taken into consideration.
Experiments of this section are launched on the test database ND CrossSensor Iris 2013 Dataset-LG4000. 
The protocols are the same as in Section~\ref{exp:occ_iris}. 
The percentage of occluded region is set as 50\%. 

%-----------------------------------------------------------------------------------------------

\subsubsection{Deep Learning Framework}
Two deep learning frameworks are utilized: the proposed method and the backbone model. In the situation without the masking strategy, the occluded samples are sent to networks directly. In the situation with masking strategy, occluded areas of samples are masked by ground truth masks.

\begin{figure}[h]
\begin{center}
\includegraphics[width=\linewidth]{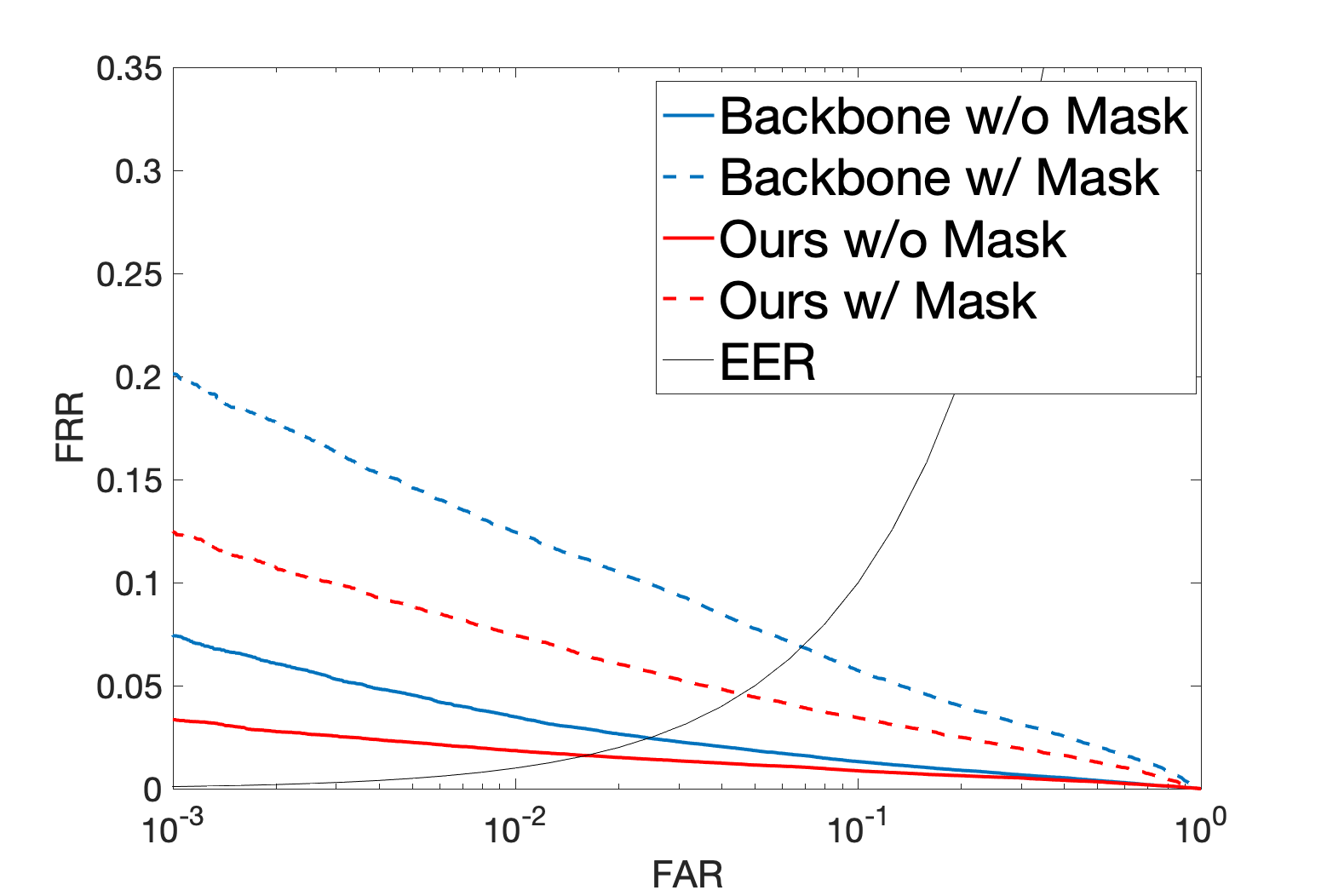}
\end{center}
\setlength{\abovecaptionskip}{0pt}
\setlength{\belowcaptionskip}{0pt}
   \caption{Comparisons to masking strategy. The percentage of occluded area is 50\%. The results show that the performance of deep learning-based methods degrades because of the distraction caused by masks.}
\label{fig:maskDeep}
\end{figure}

The results are shown in Fig.~\ref{fig:maskDeep}. The performance with the adopted masking strategy is worse than that without the masking strategy for both frameworks, as observed in Fig.~\ref{fig:maskDeep}. 
The results indicate that masking occluded regions out before feature extraction can be harmful to deep learning frameworks. The masks distract the networks by introducing incorrect contextual information.
%The reasons may be that the masks disturb the feature extraction. 
%For example, the masks of two images from different class are the same in shape and location, the inter-class distance will be decreased because of the masks. 
%The results indicate that covering occluded areas of input images by masks is harmful to deep learning framework. 

%-----------------------------------------------------------------------------------------------

\subsubsection{Handcrafted Approach}
The 2D Gabor filter is a common tool for context analysis, and it has been proven to be effective for iris recognition~\cite{Daugman1993High}. Dynamic graph matching and the masking strategy are both applied to 2D Gabor features for comparison.

For the purpose of applying dynamic graph matching to 2D Gabor features, FGs are established based on the 2D Gabor features.
Binary features extracted by Gabor filters are reorganized as a tensor $\mathcal{F} \in \mathbb{R}^{C\times H\times W}$, where $H\times W$ is the spatial size of $\mathcal{F}$, and $C$ is the number of channels, which depends on the parameters of Gabor filters. FGs are established from $\mathcal{F}$ in two steps: 
%Modified 4-th Minor: local feature $\mathcal{N}_i \in \mathbb{R}^{C\times S\times S}$ is selected from $\mathcal{F}$ as the representation of a node according to the locations of nodes, where $i \in \{ 1, 2, ..., N \}$, $N$ is the number of nodes,
1) local feature $\mathcal{N}_i \in \mathbb{R}^{C\times S\times S}$ is selected from $\mathcal{F}$ as the representation of a node according to the locations of nodes, where $i \in \{ 1, 2, ..., N \}$, $N$ is the number of nodes, $S$ is the scale of the local features, and the locations of nodes (normalized by the spatial size of $\mathcal{F}$) yielded by original MS-DGR are used for selection. The locations of nodes are the spatial centers of local features.
2) an adjacent matrix is yielded according to Eq. \ref{equ:gaussian}, where $R = S$.
Dynamic graph matching described in Section~\ref{med:dgm} is applied to similarity calculation.

In the experiment, 40 Gabor filters are selected for feature extraction, and 3 channels (real part, imaginary part and energy) are extracted from each filter. Only the medium-scale locations yielded by the original MS-DGR are utilized in these experiments, and the scale of local feature $S = 9$.

\begin{figure}[h]
\begin{center}
\includegraphics[width=\linewidth]{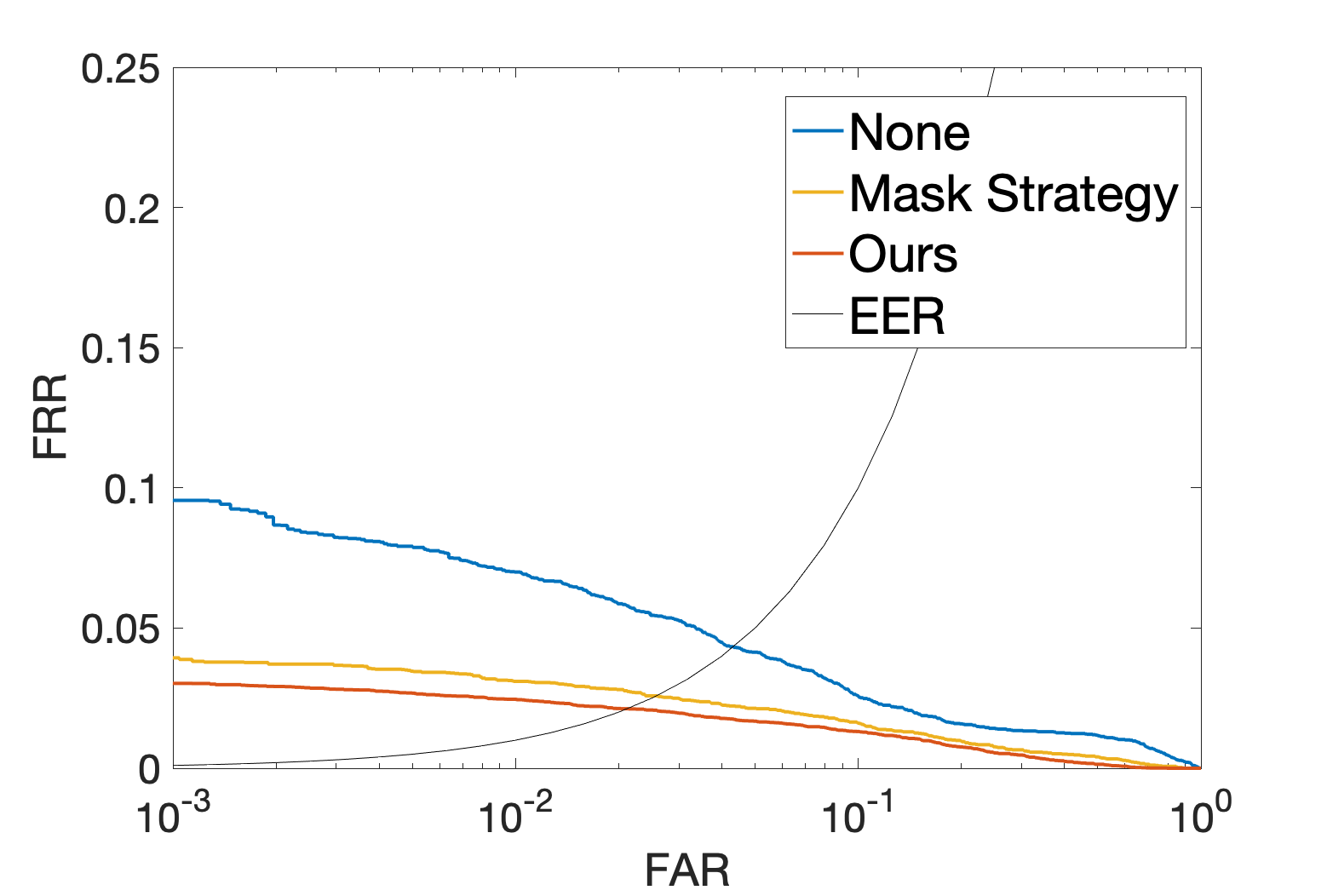}
\end{center}
\setlength{\abovecaptionskip}{0pt}
\setlength{\belowcaptionskip}{0pt}
   \caption{Comparisons to masking strategy. The percentage of occluded area is 50\%. The results show that the proposed framework is comparable with the masking strategy. This framework even functions slightly better than the masking strategy.}
\label{fig:maskGabor}
\end{figure}

The results are shown in Fig.~\ref{fig:maskGabor}. Dynamic graph matching and masking strategy are both better than bare features. The proposed framework is better than the mask strategy. The results indicate that the proposed framework not only is effective for deep learning framework but also works well in handcrafted cases.% It is a unified approach for handling occlusion.

%-----------------------------------------------------------------------------------------------
%-----------------------------------------------------------------------------------------------

\subsection{Experiments on Different Scales}

Experiments investigating the functionality of MS-DGR are launched to explore the characteristics of the FGs at different scales. For the test of the graph representation of a single scale, the other two scales are removed from the framework. All three scales are tested on the ND-LG4000 database. 

\begin{figure}[h]
\begin{center}
\includegraphics[width=\linewidth]{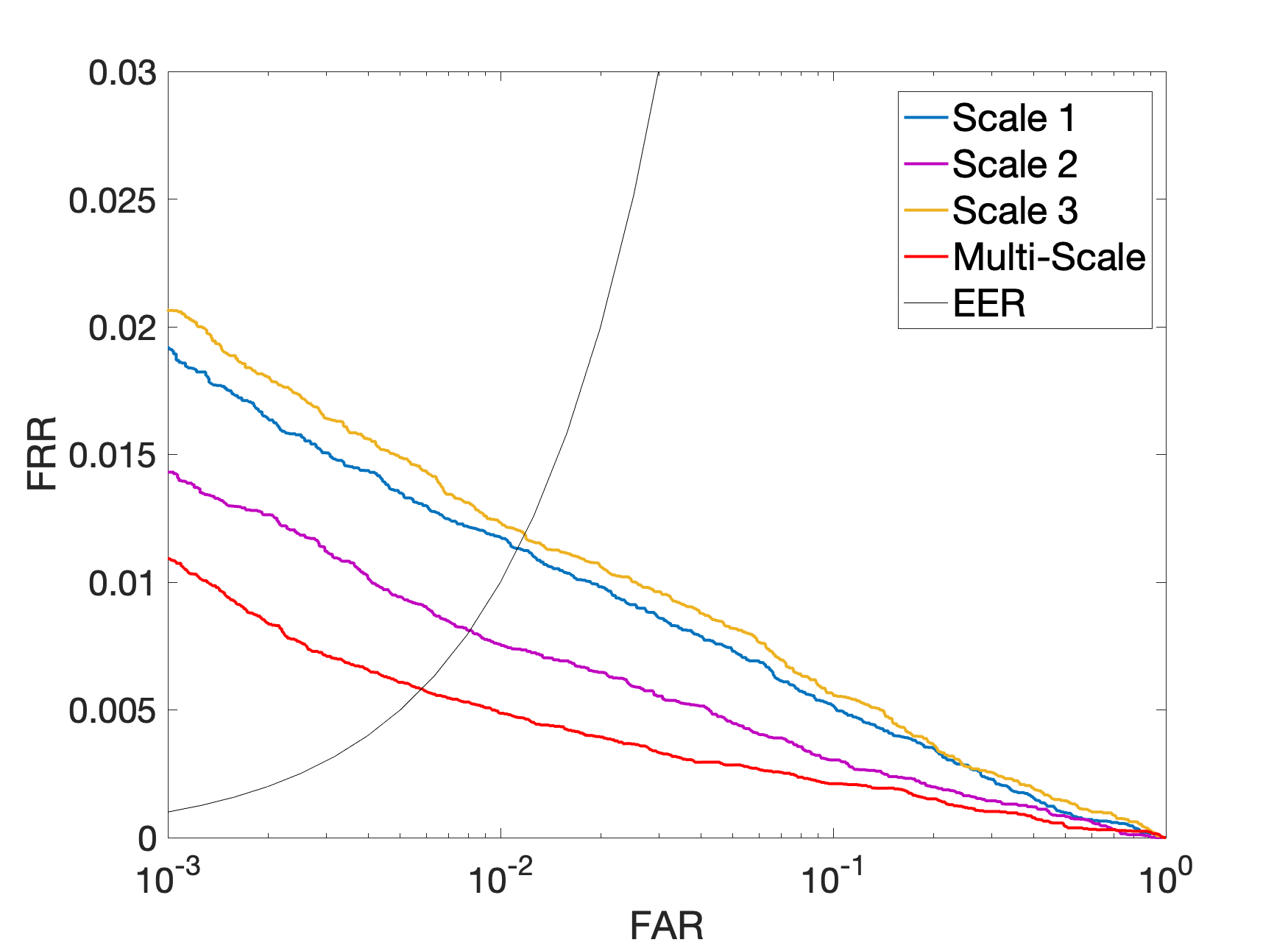}
\end{center}
   \caption{Performance of different scales and multiscale. The FGs of different scales represent the contextual information and structural information of the iris at different scales. The performance of the multiscale representation is better than all single-scale representations since multiscale representations adaptively fuse patterns of different scales.}
\label{fig:scale}
\end{figure}

The results are shown in Fig.~\ref{fig:scale} and Tab.~\ref{tab:scale}, where Scale 1 means that the small-scale FG is kept and the other two scales are removed, Scale 2 retains the medium-scale FG, and Scale 3 retains the largest-scale FG.

Large-scale patterns, such as crypts, are encoded by large-scale FGs. Patterns of small scale, such as fine furrows, are encoded by small-scale FGs.
The FGs of different scales represent the contextual information and structural information of the iris at different scales. 

%The performance of Scale 2 is the best of single-scale representations as we can observe from the results. The reasons may be that the discrimination of each node of Scale 1 is not enough which limited the capability of the nodes of graphical representations, and the spatial resolution of Scale 3 is lower than Scale 2 which limited the capability of the adjacent matrix of graphical representations. 

%Modified 4-th Minor:  multiscale representations adaptively fuse patterns of different scales.
The performance of the multiscale representation is better than all single-scale representations, because multiscale representations adaptively fuse patterns of different scales.

%Modified 4-th Minor:  multiscale representations adaptively fuse patterns of different scales

\begin{table}
\begin{center}
\caption{FRR and EER of different scales and multiscale.}
\label{tab:scale}

\setlength{\tabcolsep}{5mm}
{
\begin{tabular}{c|c|c}
\hline
& FRR@FAR=0.1\% &100\%-EER \\
\hline
Scale 1& 1.92\% & 1.13\%  \\
\hline
Scale 2& 1.43\% &0.81\%  \\
\hline
Scale 3& 2.06\% & 1.19\%  \\
\hline
Multiscale& \bf{1.09\%} & \bf{0.58\%}  \\
\hline
\end{tabular}}
\end{center}
\end{table}

%-----------------------------------------------------------------------------------------------
%-----------------------------------------------------------------------------------------------

\subsection{Experiments on the Number of Nodes}

The number of nodes of each FG is an important hyperparameter of the proposed framework. In this section, we explore the influence of different numbers of nodes. The experiments in this section are launched on the medium scale, and the other two scales of FGs are removed from the framework to clearly show the influence of different numbers of nodes.

\begin{figure}[h]
\begin{center}
\includegraphics[width=\linewidth]{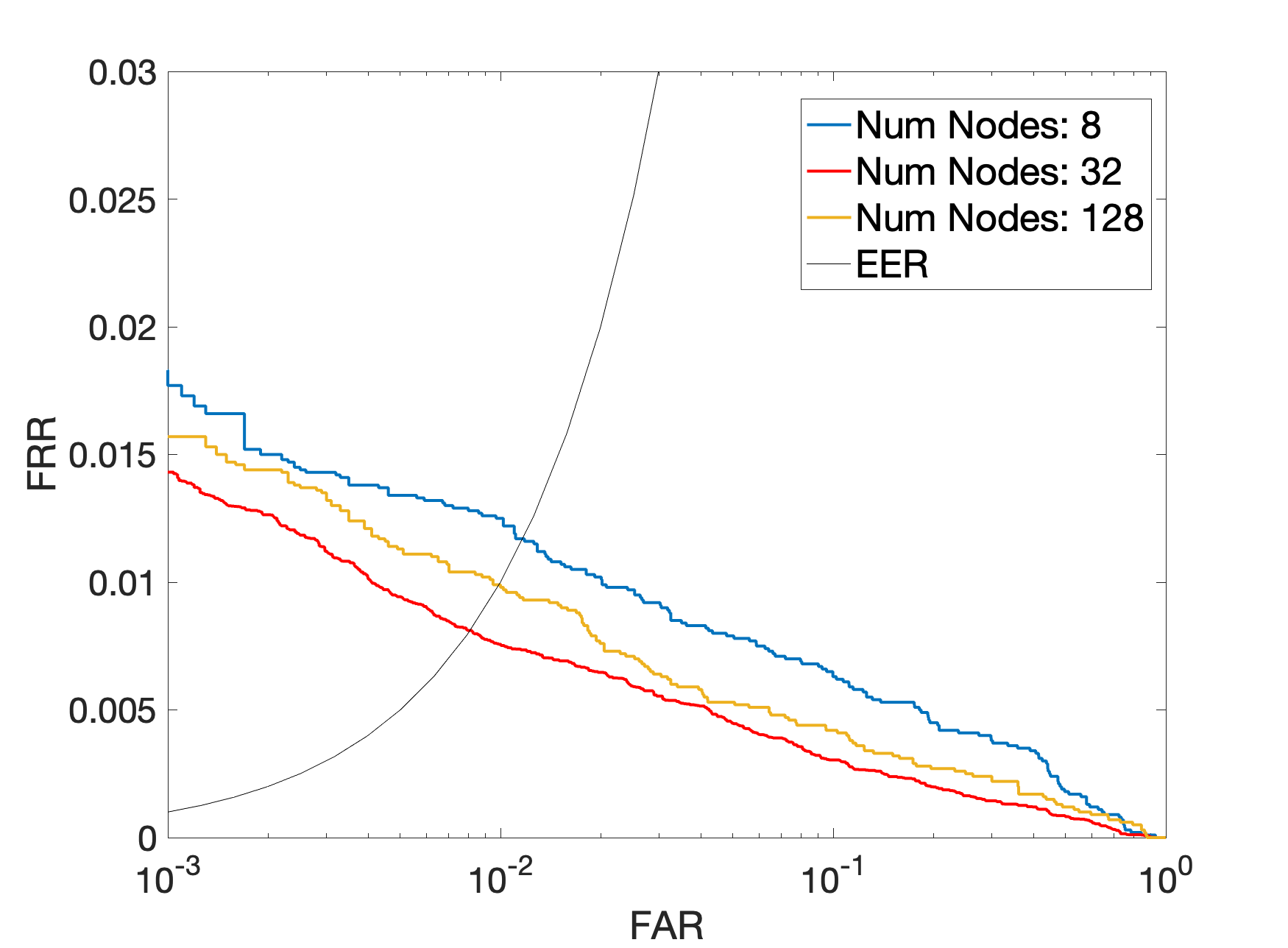}
\end{center}
   \caption{Performance of different numbers of nodes (medium scale). When the number of nodes is too few, the capability of FG is not sufficient. In contrast, when there are too many nodes, FG exhibits overfitting in some ways.}
\label{fig:node}
\end{figure}

Experiments are launched on the ND-LG4000 database. The results are shown in Fig.~\ref{fig:node} and Tab.~\ref{tab:node}. The dimensions of features are also shown in Tab.~\ref{tab:node}. The performance of 32 nodes is superior to those with 8 nodes and 128 nodes. 

%Modified 4-th Minor:  FG is a container of the local patterns and spatial relationships of a certain scale
FG is a container of the local patterns and spatial relationships of a certain scale. If the number of nodes is insufficient, the capability of FG is not sufficient to express the local texture and spatial relationships. In contrast, if there are too many nodes, FG exhibits overfitting on the local patterns and spatial relationships in some ways.
%The reason may be that some distractive information is included when there are too many nodes and the capability of representations is limited when there are too few nodes.

\begin{table}
\begin{center}
\caption{Performance of different numbers of nodes at the medium scale.}
\label{tab:node}
\setlength{\tabcolsep}{2.mm}
{
\begin{tabular}{c|c|c|c}
\hline
Num of Nodes& FRR@FAR=0.1\% &EER&Feature Dimension \\
\hline
8& 1.77\% & 1.17\%& 512 \\
\hline
32& \bf{1.43\%} &\bf{0.81\%}&1280  \\
\hline
128& 1.57\% & 0.99\%&4352  \\
\hline
\end{tabular}}
\end{center}
\end{table}

%-----------------------------------------------------------------------------------------------
%-----------------------------------------------------------------------------------------------
\subsection{Ablation Experiments}

Ablation experiments on the graph module are conducted to evaluate the effect of the graph module in this section. In addition, ablation experiments on SE-GAT are conducted to evaluate the effects of residual structure and SE layer.

%-----------------------------------------------------------------------------------------------
\subsubsection{The Effect of the Graph Module}

% Modified 4-th Minor: non-occluded, instances
Two protocols are taken into consideration to evaluate the graph module: with and without the graph module. In the protocol without the graph module, all of the graph blocks are removed from the proposed framework, which makes the framework degenerate into CNNs. In the protocol with the graph module, graph blocks remain. The experiments are launched on the ND-LG4000 database, and both non-occluded and occluded instances are taken into consideration.

\begin{figure}[h]
\begin{center}
\includegraphics[width=0.95\linewidth]{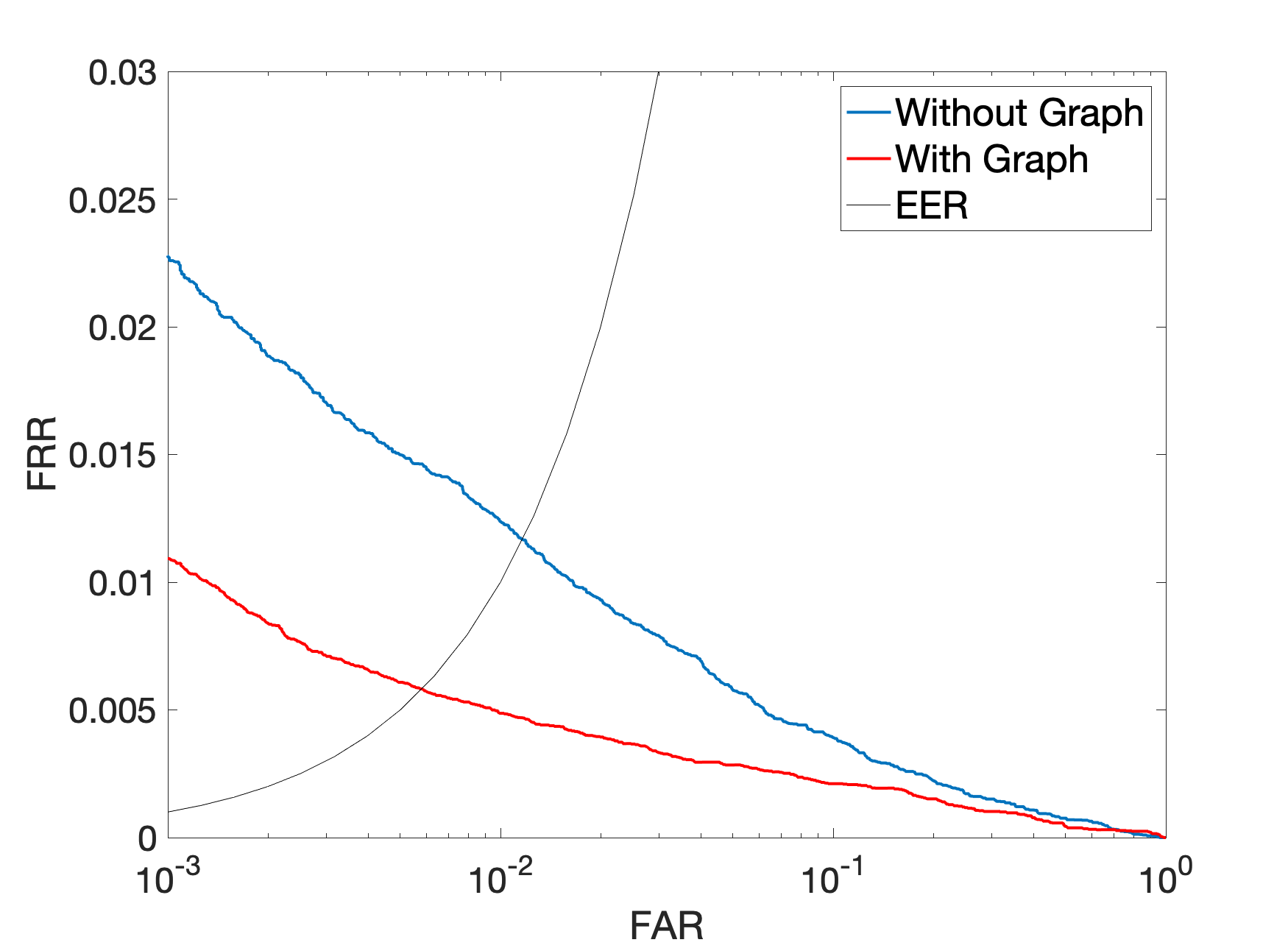}
\end{center}
   \caption{Results of the ablation study on ND-LG4000.}
\label{fig:abl_l4}
\end{figure}

\begin{figure}[h]
\begin{center}
\includegraphics[width=0.95\linewidth]{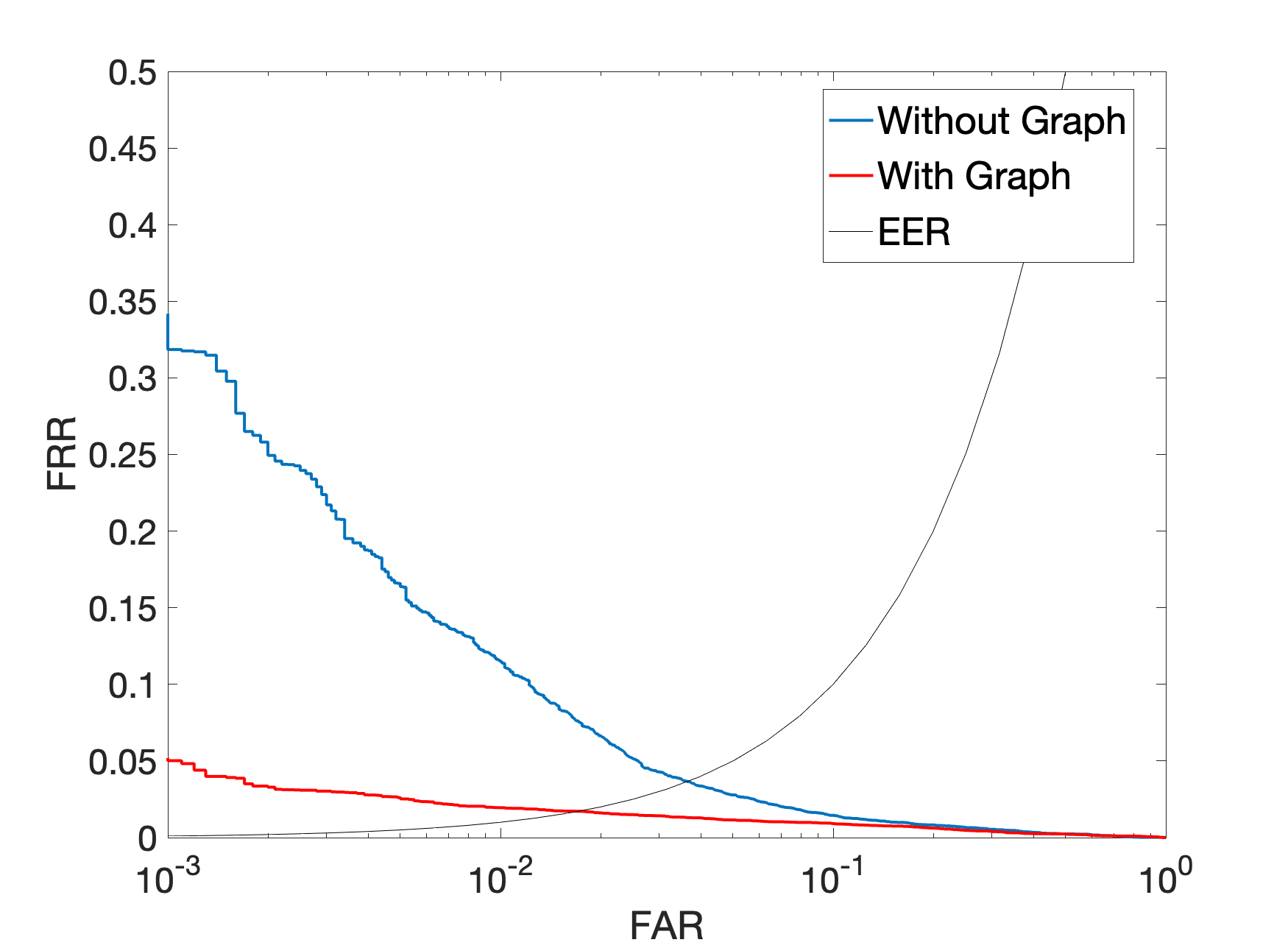}
\end{center}
   \caption{Results of the ablation study on occluded instances.}
\label{fig:abl_occ5}
\end{figure}

The results of the non-occluded instances are shown in Fig.~\ref{fig:abl_l4} and Tab.~\ref{tab:abl_l4}. The results of the occluded instances are shown in Fig.~\ref{fig:abl_occ5} and Tab.~\ref{tab:abl_occ5}. The performance gap between the protocols with and without the graph module is obvious, which demonstrates the effectiveness of the graph module.
%Modified 4-th Minor: effective
The gap in the case of occluded situations is larger, which indicates that the graph module is effective for occluded biometric scenarios.

\begin{table}
\begin{center}
\caption{FRR and EER of the ablation study on ND-LG4000.}
\label{tab:abl_l4}
\setlength{\tabcolsep}{5.mm}
{
\begin{tabular}{c|c|c}
\hline
& FRR@FAR=0.1\% &EER\\
\hline
without Graph & 2.27\% & 1.17\% \\
\hline
with Graph & \bf{1.09\%} &\bf{0.58\%}  \\
\hline
\end{tabular}}
\end{center}
\end{table}

\begin{table}
\begin{center}
\caption{FRR and EER of the ablation study on occluded instances.}
\label{tab:abl_occ5}
\setlength{\tabcolsep}{5.mm}
{
\begin{tabular}{c|c|c}
\hline
& FRR@FAR=0.1\% &EER\\
\hline
without Graph & 31.85\% & 3.66\% \\
\hline
with Graph & \bf{5.09\%} &\bf{1.72\%}  \\
\hline
\end{tabular}}
\end{center}
\end{table}

%-----------------------------------------------------------------------------------------------

\begin{figure}[t]
\begin{center}
\includegraphics[width=0.95\linewidth]{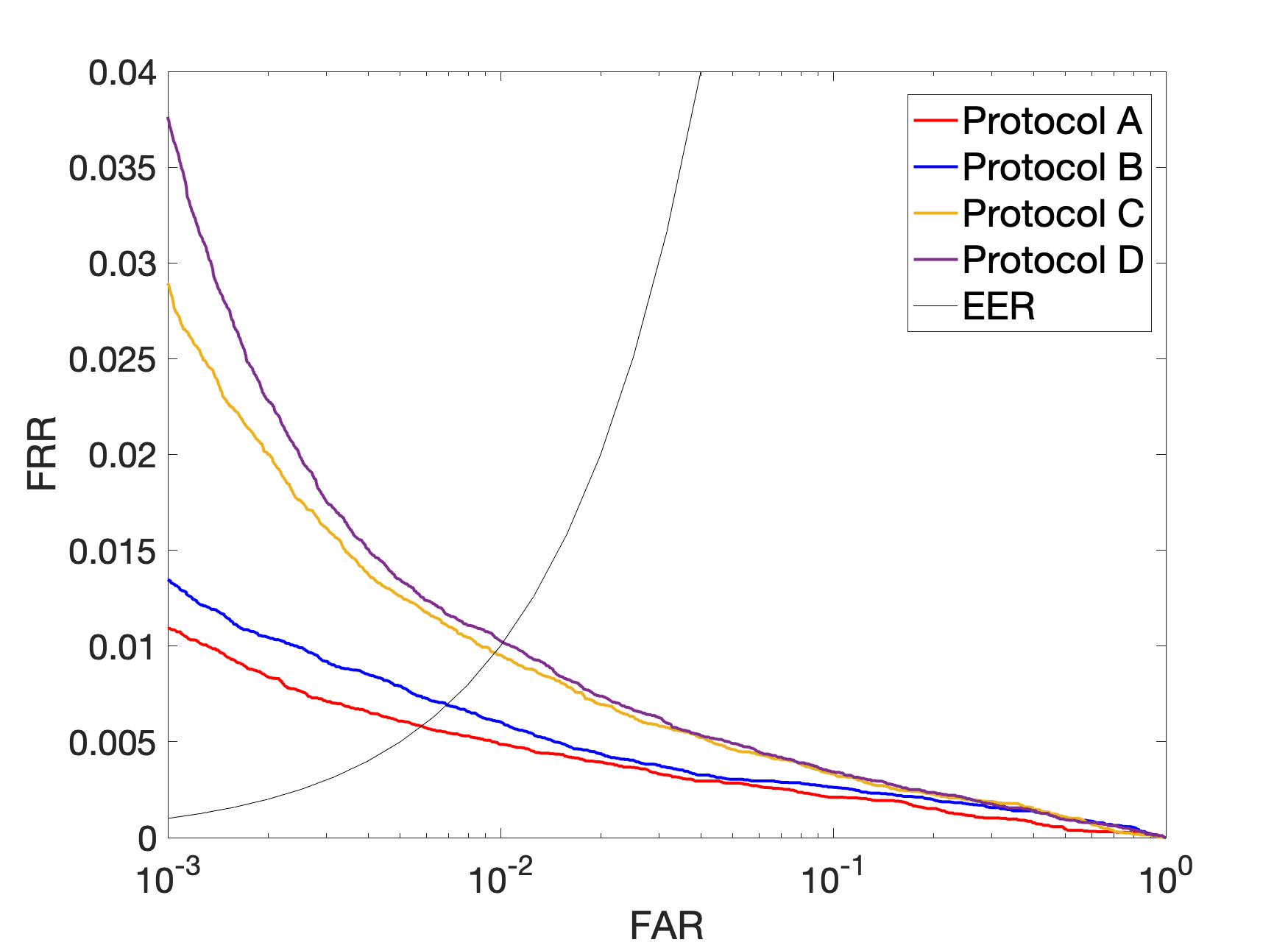}
\end{center}
   \caption{Results of the ablation study on SE-GAT. The results show that residual structure and SE layer are both beneficial, and the contribution of the residual structure is more distinct.}
\label{fig:abl_segat}
\end{figure}

\begin{table}[t]
\begin{center}
\caption{Results of the ablation study of SE-GAT.}
\label{tab:abl_segat}
\setlength{\tabcolsep}{1mm}
{
\begin{tabular}{|c|c|c|c|c|}
\hline
& Residual & SE Layer & FRR@FAR=0.1\% & EER \\
\hline
Protocol A & \checkmark & \checkmark & \bf{1.09\%} & \bf{0.58\%}  \\

Protocol B & \checkmark & $\times$ & 1.35\% & 0.67\%  \\

Protocol C & $\times$ & \checkmark & 2.89\% & 0.95\%  \\

Protocol D & $\times$ & $\times$ & 3.76\% & 1.01\%  \\
\hline
\end{tabular}}
\vspace{-0.3cm}
\end{center}
\end{table}

\subsubsection{Ablation Study on SE-GAT}

Ablation experiments on SE-GAT are conducted to evaluate the effects of residual structure and SE layer. There are four protocols: protocol A with both strategies adopted, protocol B with only a residual structure adopted, protocol C with only the SE layer adopted, and protocol D, which abandons both strategies. The results are shown in  Fig.~\ref{fig:abl_segat} and Tab.~\ref{tab:abl_segat}.

The results show that both strategies contribute to the performance of the proposed framework. The residual structure is useful to reduce the false rejection rate, which means narrowing the intraclass variance.

%------------------------------------------------------------------------------
%------------------------------------------------------------------------------
%----------------------------

\section{Conclusion}
\label{sec:Conclusion}
In this paper, a novel deep learning framework called MS-DGR is proposed for handling occlusion problems in biometric recognition.
%
%Modified 4-th Minor: The effectiveness of the proposed framework is verified on two biometric modalities, i.e. iris and face.
The effectiveness of the proposed framework is verified on two biometric modalities, i.e. iris and face.
%Two biometric modalities, iris and face, are adopted to verify the effectiveness of the proposed framework.

%Dynamic graphs are adopted to overcome the occlusion situations. A novel deep graph model is proposed for processing of the FG. Multi-scale strategy is adopted to extract and fuse the information of different scales. 

%Modified 4-th Minor: no longer a prerequisite
The proposed framework provides a novel strategy for occluded biometric recognition. Prior knowledge about the occluded region is no longer a prerequisite, and the inference is more illustrative and reasonable because of the proposed graph representation and dynamic graph matching. 
%Modified 4-th Minor: selected adaptively by the proposed framework
Different from most existing methods, the features are not selected according to mask but selected adaptively by the proposed framework.
%The proposed framework provides a novel strategy for occluded biometrics recognition, which is more effective than masking strategy in scenarios of deep learning. Performances are better and the costly segmentation labeling is not needed. 
The framework is well designed for both deep learning and handcraft methods.%the architecture of neural networks, and the components can be updated with the development of graphical model and machine learning methods.

The main idea of this paper is straightforward, and we believe that there is much room for improvement. Further explorations of the proposed framework could focus on better methods for graph generation, multi-scale fusion, and dynamic graph matching.

% or


%\appendices

%\section{}

\ifCLASSOPTIONcompsoc
  \section*{Acknowledgments}
\else
  \section*{Acknowledgment}
\fi

The authors would like to thank the reviewers for their valuable comments and advices. 
This work is funded by National Natural Science Foundation of China (Grant No. 62276263, 62006225, 62071468) in part by the National Key Research and Development Program of China under Grant 2022YFC3310400, and supported by the Strategic Priority Research Program of Chinese Academy of Sciences (Grant No. XDA27040700).
%

%%

%This work is jointly supported by National Key Research and Development Program of China under Grant 2022YFC3310400, the National Natural Science Foundation of China (Grant No. 62276025, 62276263, 62006225, 62071468) and Shenzhen Technology Plan Program (KQTD20170331093217368).

\ifCLASSOPTIONcaptionsoff
  \newpage
\fi



%\bibliographystyle{IEEEtran}
\bibliographystyle{unsrt}
\end{document}